\documentclass{article}

\PassOptionsToPackage{numbers, compress}{natbib}

    \usepackage[final]{neurips_2024}

\usepackage[numbers]{natbib}
\usepackage[utf8]{inputenc} 
\usepackage[T1]{fontenc}    
\usepackage{hyperref}       
\usepackage{url}            
\usepackage{booktabs}       
\usepackage{amsfonts}       
\usepackage{nicefrac}       
\usepackage{microtype}      
\usepackage[dvipsnames]{xcolor}        

\definecolor{mycitecolor}{HTML}{3498DC}
\definecolor{mylinkcolor}{HTML}{E74D3B}
\definecolor{myurlcolor}{HTML}{980000}
\definecolor{mydarkgreen}{rgb}{0,0.6,0}
\definecolor{myorange}{HTML}{E7730D}
\definecolor{myblue}{HTML}{4594c1}

\definecolor{AAGray}{HTML}{999999}
\definecolor{AABlue}{HTML}{6db0ce}
\definecolor{AARed}{HTML}{ce471a}

\usepackage{graphicx}
\usepackage{wrapfig}
\usepackage{amsmath}
\usepackage{subcaption}
\usepackage{tcolorbox}
\usepackage{enumitem}
\usepackage{wrapfig}
\usepackage{tocloft}
\newcommand{\hypothesis}[2]{%
\begin{tcolorbox}[colback=OliveGreen!10!white,leftrule=2.5mm,size=title]
\textbf{#1}: #2
\end{tcolorbox}
}

\definecolor{darkgreen}{HTML}{C00000}

\title{Is Mamba Compatible with Trajectory
Optimization in Offline Reinforcement Learning}

\author{ 
Yang Dai$^1$
\quad
Oubo Ma$^2$
\quad
Longfei Zhang$^1$
\quad
Xingxing Liang$^1$\footnotemark[1] 
\\
\vspace{0.2cm}
\textbf{
Shengchao Hu$^3$
\quad
Mengzhu Wang$^4$
\quad
Shouling Ji$^2$
\quad
Jincai Huang$^1$
\quad
Li Shen$^5\thanks{Corresponding authors: Li Shen and Xingxing Liang.}$
}\\
\small{$^1$Laboratory for Big Data and Decision, National University of Defense Technology}\\
\small{$^2$Zhejiang University}
\quad
\small{$^3$Shanghai Jiao Tong University
}\\
\quad 
\small{$^4$Hebei University of Technology}
\quad
\small$^5${Shenzhen Campus of Sun Yat-sen University}\\
{\tt\small
\{daiyang2000, zhanglongfei, liangxingxing, huangjincai\}@nudt.edu.cn}\\
{\tt\small \{mob, sji\}@zju.edu.cn; charles-hu@sjtu.edu.cn; \{dreamkily,mathshenli\}@gmail.com}
}
\begin{document}

\maketitle
\begin{abstract}
Transformer-based trajectory optimization methods have demonstrated exceptional performance in offline Reinforcement Learning (offline RL). Yet, it poses challenges due to substantial parameter size and limited scalability, which is particularly critical in sequential decision-making scenarios where resources are constrained such as in robots and drones with limited computational power. Mamba, a promising new linear-time sequence model, offers performance on par with transformers while delivering substantially fewer parameters on long sequences. As it remains unclear whether Mamba is compatible with trajectory optimization, this work aims to conduct comprehensive experiments to explore the potential of Decision Mamba (dubbed DeMa) in offline RL from the aspect of data structures and essential components with the following insights: (1) Long sequences impose a significant computational burden without contributing to performance improvements since DeMa's focus on sequences diminishes approximately exponentially. Consequently, we introduce a Transformer-like DeMa as opposed to an RNN-like DeMa. (2) For the components of DeMa, we identify the hidden attention mechanism as a critical factor in its success, which can also work well with other residual structures and does not require position embedding. Extensive evaluations demonstrate that our specially designed DeMa is compatible with trajectory optimization and surpasses previous methods, outperforming Decision Transformer (DT) with higher performance while using 30\% fewer parameters in Atari, and exceeding DT with only a quarter of the parameters in MuJoCo.
\end{abstract}
\setcounter{footnote}{0} 
\addtocontents{toc}{\protect\setcounter{tocdepth}{0}}
\section{Introduction}

Offline Reinforcement Learning (Offline RL)~\cite{levine2020offline} has gained significant attention due to its ability to learn strategies without interacting with the environment, which is particularly beneficial in situations where real-time interaction is expensive or risky~\cite{kiran2021deep, brandfonbrener2021offline, afsar2022reinforcement}. 
With a static dataset, offline RL can be implemented through three distinct learning methods~\cite{prudencio2023survey}: (1) model-based algorithm~\cite{kidambi2020morel, lu2021revisiting, he2023survey}, (2) model-free algorithm~\cite{swazinna2022comparing,haarnoja2018soft, fujimoto2019off}, (3) trajectory optimization\cite{chen2021decisiontransformer, dc,hu2023graph, tt,hu2024transforming}. The first two methods require long-term credit assignment through the Bellman equation, leading to the "deadly triad" problem known to destabilize RL~\cite{sutton2018reinforcement}. 
In contrast, trajectory optimization methods treat RL problems as sequence modeling problems to get better performance and generalization~\cite{chen2021decisiontransformer}. Most trajectory optimization methods rely on transformers, which perform credit assignment directly through the attention mechanism. By leveraging the powerful modeling capabilities of transformers, these methods outperform other offline RL algorithms~\cite{radford2018improving, radford2019language, brown2020language}.

The transformer attention mechanism \cite{vaswani2017attention}, which allows the model to focus on the important part of the input sequence~\cite{chaudhari2021attentive},  has several downsides. The computational demands of the attention mechanism escalate quadratically with the input length, posing a significant constraint on its scalability~\cite{shen2021efficient, katharopoulos2020transformers,dao2023flashattention}. Moreover, some studies \cite{tay2021synthesizer, yu2022metaformer} suggest that the attention mechanism may not be the primary factor contributing to the effectiveness of transformers. This notion is also supported in offline RL, where~\cite{dc} discovers that the attention mechanism of Decision Transformer (DT) does not capture local associations effectively, rendering it unsuitable for RL. Given these limitations, we are led to ponder if a more efficient mechanism with fewer parameters and greater scalability exists for offline RL.
Recently, a series of state space models (SSMs)~\cite{hamilton1994state}, particularly Mamba~\cite{gu2023mamba}, have been proposed as potential solutions with the ability to scale linearly concerning the sequence length. In particular, Mamba introduces a selective hidden attention mechanism~\cite{hidden_attention} for content-based reasoning and employs parallel scan to enhance computational efficiency, resulting in two approaches to employing Mamba in offline RL. The first is the Transformer-like Mamba, a direct substitution of the transformer~\cite{li2024videomamba,liu2024vmamba,zhu2024visionmamba} while the other is the RNN-like Mamba~\cite{ma2024umamba}, achieving an inference speed with constant time complexity. 

Few studies have explored the application of SSMs in offline RL, though they perform well in model-based algorithms~\cite{samsami2024mastering,deng2024facing} and in-context RL learning~\cite{lu2024structured}.
Mamba is tailored for memory-required long-sequence tasks, whereas trajectory optimization methods typically utilize short segments during training and inference, as most RL tasks are modeled as Markov Decision Processes (MDPs), i.e. past information may not influence current decisions. 
Furthermore, due to the lack of a comprehensive investigation of the key component of Mamba, a question has arisen:

\begin{center}
\textit{\textbf{Whether Mamba is compatible with trajectory optimization?}}
\end{center}

In this work, we aim to undertake a thorough investigation and in-depth analysis to explore this question. Specifically, we focus on the data structures and the essential components in trajectory optimization. The extensive experiments provide strong support for the following key findings. (1) We explore the data structures with an analysis of sequence length and concatenating type. The former reveals that long input sequences present computational challenges without enhancing performance due to the hidden attention scores of DeMa evincing an exponential decay pattern. As a result, we opt for the Transformer-like DeMa as opposed to the RNN-like DeMa for efficiency and effectiveness. The latter finds concatenating in the temporal dimension is better for the Transformer-like DeMa. (2) The hidden attention mechanism plays a pivotal role in DeMa's effectiveness and is compatible with the transformer's post up-projection residual structure~\cite{beck2024xlstm}, enabling it to replace the attention layer directly and eliminating the need for position embedding. Extensive evaluations show that with a higher average score and nearly 30\% fewer parameters, DeMa significantly outperforms DT in eight Atari games. Furthermore, in nine MuJoCo tasks, DeMa's performance not only exceeds that of DT but does so with only one-fourth of the parameters, highlighting remarkable improvements in both performance efficiency and model compactness.

In the end, our main contributions can be summarized as follows:
\begin{enumerate}[itemsep=1pt,parsep=1pt,topsep=0pt,partopsep=0pt]
    \item We find the Transformer-like DeMa surpasses the RNN-like DeMa in both efficiency and effectiveness for trajectory optimization. Extensive experiments on sequence length and concatenating type show the impact of the input data, which guides the design of DeMa.
    \item Through various ablation experiments, we discover that the hidden attention mechanism is the core component in DeMa and does not require position embedding. This finding enhances the effectiveness and efficiency of our Transformer-like DeMa.
    \item With state-of-the-art performance on both MuJoCo and Atari, our Transformer-like DeMa significantly addresses the challenges posed by transformer-based trajectory optimization methods, particularly the issues of large parameter sizes and limited scalability.
\end{enumerate}
\section{Related Work}
\paragraph{Offline RL.} Offline RL is a data-driven RL paradigm in which the agent learns solely from a pre-collected dataset rather than through interaction with the environment~\cite{hu2024transforming}. Distribution shifts~\cite{fujimoto2019off} can severely impact performance when RL algorithms are deployed directly in offline environments, leading to significant degradation. To mitigate this problem, several methods have been introduced, which the study~\cite{prudencio2023survey} categorizes into three primary approaches: (1) learning a dynamics model to generate additional training data (model-based algorithm)~\cite{kidambi2020morel, yu2021combo}, (2) learning a policy through a model-free approach by constraining unseen actions or incorporating pessimism into the value function (model-free algorithm)~\cite{haarnoja2018soft, fujimoto2019off, liuadaptive}, and (3) trajectory optimization~\cite{chen2021decisiontransformer, tt}.
The method of trajectory optimization is usually based on a causal transformer model and converts an RL problem to a sequence modeling problem~\cite{dc}. It performs credit assignment directly through the attention mechanism in contrast to Bellman backups, thus modeling a wide distribution of behaviors, enabling better generalization and transfer~\cite{chen2021decisiontransformer}.

\paragraph{Sequence Modeling in Offline RL.}
Following DT~\cite{chen2021decisiontransformer} and Trajectory Transformer (TT)~\cite{tt}, there has been an increasing trend in employing advanced sequence-to-sequence model to solve RL tasks~\cite{hu2023graph, prompt_tuning_dt, QT, HarmoDT, CommFormer, hu2023instructed, huang2024context}.\footnote{For detailed insights, one may refer to the relevant comprehensive reviews~\cite{hu2024transforming, li2023survey}.}  Unfortunately, these improvements are usually transformer-based and hence suffer from the common dilemma of the attention mechanism, i.e. over-parameterization and inability to scale to long sequence tasks. What's more, Emmons et al.~\cite{emmons2021rvs} find that simply maximizing likelihood with a two-layer feedforward MLP is close to the results of substantially more complex methods based on sequence modeling with Transformers. Similarly, Lawson et al.~\cite{lawson2023merging} find that replacing the attention parameters with those learned in other environments has a minimal impact on the performance. Besides, Decision ConvFormer (DC)~\cite{dc} indicates that substituting the attention layers with learnable parameters can lead to improved outcomes. 
These observations suggest significant redundancy in the Transformer architecture, highlighting the potential to explore lighter and more scalable networks for implementation in offline RL.
Building on this, the Structure SSM (S4)~\cite{gu2021s4} has emerged as a promising alternative. 
Studies \cite{samsami2024mastering} and \cite{deng2024facing} use S4 in model-based RL, outperforming traditional Transformer and RNN approaches. The capabilities of S4 and Mamba are further demonstrated by~\cite{lu2024structured, huang2024decision}, which points to their speed and effectiveness in in-context RL tasks.

The most related work to ours is Decision S4 (DS4)~\cite{david2022ds4} and Decision Mamba (DMamba)~\cite{ota2024decisionmamba}, where the former uses an RNN-like S4 for inference, and the latter replaces the attention mechanism with Mamba directly. In contrast, our work finds that Transformer-like DeMa outperforms RNN-like DeMa as the long sequences impose a significant computational burden on Mamba without contributing to performance improvements. What's more, DMamba simply substitutes Mamba for the attention block rather than the transformer block while our investigation shows the key component is the hidden attention mechanism, which eliminates the need for position embedding and hence achieves better performance with fewer parameters.

\section{Preliminaries}

In this section, we present several necessary preliminaries and terminologies of offline RL, trajectory optimization, state space model, and hidden attention in Mamba. 

\subsection{Offline RL with Trajectory Optimization} 

Given a static dataset of transitions $\tau=\{(s_t,a_t,s_{t+1},r_t)_i\}$, where $i$ presents the timestep of a transition in the dataset. The states and actions are generated by the behavior policy $(s_t,a_t)\sim d^{\pi_\beta}(\cdot)$, while the next states and rewards are determined by the unknown transition dynamics $p(s^{\prime},r|s, a)$. The goal of offline RL is to find an approximate policy $\pi(a|\cdot)$ that maximizes expected return $\mathbb{E}[\sum_{t=0}^{T}r_{t}]$, where $T$ represents the time step at which the episode terminates. Due to the lack of interaction with the environment, trajectory optimization methods transform the goal into minimizing reconstruction loss, i.e. minimizing loss $\mathbb{E}_{(\hat{R},s, a)\sim\tau}[\frac{1}{T}\sum_{t=1}^{T}\mathcal{L}_{\mathrm{MSE/CE}}(\hat{a}_{t}; a_{t})]$, where $\hat{a}_{t}=\pi(\cdot|s_{t-K+1:t},\hat{R}_{t-K+1:t}, a_{t-K:t-1})$, and $\hat{R}_{t} = \sum_{t^{\prime}=t}^{T}r_{t^{\prime}}$ is the return-to-go (RTG). At test time, a target RTG $R_0$ is manually set to represent the desired performance. We input the trajectories from the last $K$ timesteps into policy $\pi$, which then generates an action for the current timestep. Subsequently, the next state and reward are received from the environment. These elements are concatenated and also input into the model.  The policy is approximated through the sequential model~\cite{chen2021decisiontransformer, janner2022planning}. However, these models typically possess a large number of parameters and struggle with handling long sequences effectively. Fortunately, this issue can be addressed by using SSMs~\cite{hamilton1994state,gu2021s4,gu2023mamba}.

\subsection{State Space Model and Mamba}
There are two approaches to utilizing Mamba in RL, which are both closely related to the modeling methods of SSM. SSM is defined by the following first-order differential equation, which maps a 1-D input signal $u(t)$ to an $N$-D latent state $h(t)$ before projecting to a 1-D output signal $y(t)$~\cite{S4},
\begin{equation}\begin{aligned}
h^{\prime}(t)=Ah(t)+Bu(t),\quad 
y(t)=Ch(t)+Du(t),
\end{aligned}\end{equation}
where $A\in\mathbb{R}^{N\times N} , B\in\mathbb{R}^{N\times 1}, C\in\mathbb{R}^{1\times N}$ and $D\in\mathbb{R}$ are trainable matrices. As $u(t)$ is typically discretized as $\{u_i\}_{i=1,2,...}$, SSM can be discretized by a step size $\Delta$. Moreover, recurrent SSM can be written as a discrete convolution. Let $h_{0}=0$ and  $D=0$, we have
\begin{equation}\begin{aligned}
y_{i}=C\bar{A}^i\bar{B}u_1+C\bar{A}^{i-1}{\bar{B}}u_2+\cdots+C\bar{A}\bar{B}u_{i-1}+C\bar{B}u_i,\quad  y=u*\bar{K},
\end{aligned}\label{eq:2}\end{equation}
where $\bar{A},\bar{B}$ is the approximation discrete of $A, B$, and $\bar{K}$ is called the SSM convolution kernel and can be represented by filter
\begin{equation}\bar{K}=(C\bar{B},C\bar{A}\bar{B},\ldots,C\bar{A}^{i}\bar{B},\ldots).\end{equation}

S4 and other time-invariant models cannot select the previous tokens to invoke from their history records. To solve this problem, Mamba merges the sequence length and batch size of the inputs, allowing the matrices $B, C$ and the step size $\Delta$ to depend on the inputs. Therefore, it is a time-varying system and cannot use the convolution view. To ensure efficient training and inference with Mamba, techniques such as parallel scanning, kernel fusion, and recomputation are employed, resulting in two types of Mamba. One type is the SSM using the recursive view, referred to as RNN-like Mamba, and the other is the SSM utilizing parallel scanning, known as Transformer-like Mamba. RNN-like Mamba is akin to DS4~\cite{david2022ds4}, wherein the complete trajectory is taken as a sample and fully inputted into the model for training. Utilizing this approach, which capitalizes on the ability to capture long-term dependencies, the inference speed can be significantly increased. During the inference process, it is sufficient to input only the current tuple ($r_{t-1}$, $a_{t-1}$, $s_t$)  in conjunction with the hidden state $h_t$. Transformer-like Mamba is a direct replacement for the transformer, where we consistently truncate the input sequences to a fixed length of $K$ before their introduction into the model throughout the training and inference phases~\cite{ota2024decisionmamba,ma2024umamba,yang2024vivim,zhang2024motionmamba}.

\subsection{Hidden Attention in Mamba}
Although the role of the self-attention mechanism in offline RL remains uncertain, it is known that this mechanism allows the model to dynamically focus on different parts of the input sequences, following the Equation~\eqref{eq:1}.
\begin{equation}
   \text{Self-Attention}(x)=\alpha V(x),\quad\alpha=\mathrm{softmax}\left(\frac{QK^\top}{\sqrt{d_k}}\right),\label{eq:1} 
\end{equation}
where $Q, K, V$ represent queries, keys, and values respectively, i.e. input sequences after three linear transformations. $d_k$ is the dimension of the keys. Similarly, current research suggests that the S6 layer in Mamba can be viewed as the hidden attention mechanism with a unique data-control linear operator~\cite{hidden_attention}. Assuming the initial condition $h_0=0$, we can obtain a formula similar to Equation~\eqref{eq:2}
\begin{equation}y_i=C_i\sum_{j=1}^i\big(\Pi_{k=j+1}^i\bar{A}_k\big)\bar{B}_jx_j,\ \ 
h_i=\sum_{j=1}^i\big(\Pi_{k=j+1}^i\bar{A}_k\big)\bar{B}_jx_j,
\label{eq:33}\end{equation}
where $\bar{A}_{i}=\exp(\Delta_{i}(A))$, $\bar{B}_{i}=\Delta_{i}(B_i)$, and $\Delta_i=\text{softplus}(S_{\Delta}(x_i))$. $B_i=S_B(x_i)$, $C_i=S_C(x_i)$, with $S_B$, $S_C$ and $S_{\Delta}$ are linear projection layers. Softplus is an elementwise function that is a smooth approximation of ReLU.

Since $\bar{A_t}$ is a diagonal matrix,~\cite{hidden_attention} simplifies the hidden matrices and gets the attention mechanism of Mamba:
\begin{equation}\begin{aligned}
\text{Hidden-Attention}&(x)=\tilde{\alpha} x,\quad\tilde{\alpha}_{i,j}\approx\tilde{Q}_i\tilde{H}_{i,j}\tilde{K}_j\\
\tilde{Q}_{i}:=S_{C}(x_{i}), \tilde{K}_{j}:=\mathrm{ReLU}(S_{\Delta}&(x_{j})S_{B}(x_{j}), \tilde{H}_{i,j}:=\exp\Big(\sum\limits_{\substack{k=j+1\\S_{\Delta}(x_{k})>0}}^{i}S_{\Delta}(x_{k})\Big)A.        
\end{aligned}\label{eq:44}\end{equation}
Therefore, we can visualize the hidden attention matrices in DeMa, thus gaining a deeper understanding of the behavior inside the model in the setting of offline RL.

\section{The Analysis of DeMa }

\begin{figure}[h]
    \centering
\includegraphics[width=.94\linewidth]{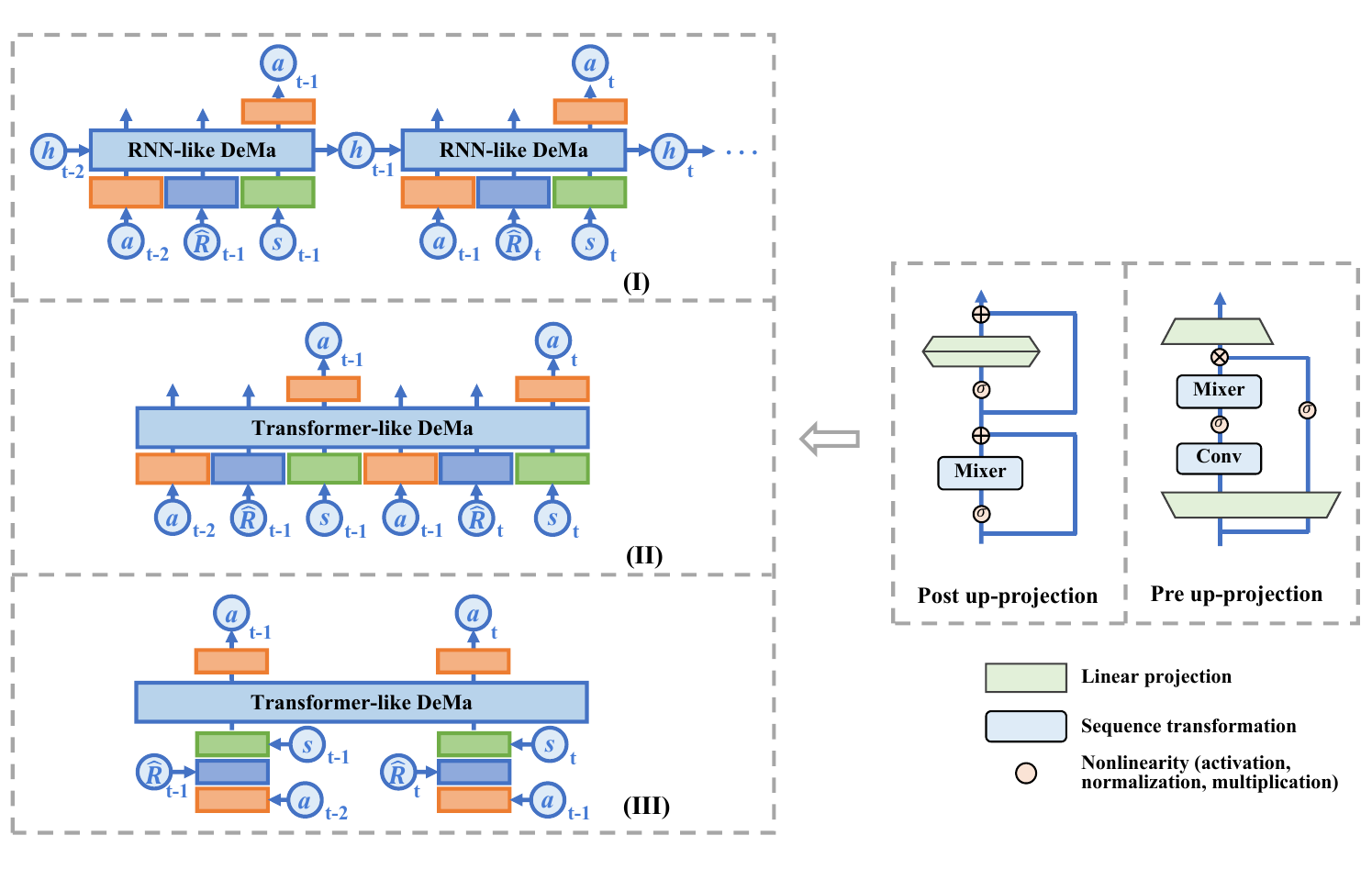}
    \caption{\small Variant design of the DeMa in trajectory optimization. In the left portion, (I) represents the RNN-like DeMa (B3LD), which requires hidden state inputs at each decision step; (II) indicates the transformer-like DeMa (B3LD); and (III) refers to the transformer-like DeMa (BL3D). The right portion illustrates that both types of these DeMa can incorporate two distinct residual structures, i.e. the post up-projection residual block and the pre up-projection residual block.}
    \label{fig:struct}
    \vspace{-0.4cm}
\end{figure}

Considering most trajectory optimization methods use short segments during both training and inference, \textit{the compatibility of Mamba with these methods remains an open question}. As shown in Figure~\ref{fig:struct}, this section presents an analysis from the perspectives of data structures and essential components. Section~\ref{sec:RNN-like} discusses the impact of data structure on trajectory optimization. Our study reveals that the RNN-like DeMa does not offer substantial benefits in terms of effectiveness or efficiency. Therefore, we investigate three critical factors: sequence length, the hidden attention mechanism, and the input concatenation types. We find that the balance between performance and efficiency highly depends on the appropriate sequence length selection. Moreover, the input concatenation method significantly influences the results, with temporal concatenation (i.e., B3LD) demonstrating its effectiveness. Section~\ref{sec:network} conducts ablation studies to identify the hidden attention mechanism as a key component of DeMa, facilitating better utilization and component replacement. Detailed experiments and additional results are in the \textbf{Appendix}. Our code is available at \url{https://github.com/AndssY/DeMa}.

\subsection{Input Data Structures}\label{sec:RNN-like}

First, we compare the RNN-like DeMa (B3LD) with the Transformer-like DeMa (B3LD)\footnote{BL3D and B3LD represent different concatenation types. Section \ref{sec:input type} gives a comprehensive explanation. Unless otherwise specified, all references to DeMa in this context refer to the B3LD type.}. The average results are shown in Table~\ref{tab:average} (with detailed results in Appendix~\ref{app:comparison}), where the performance of the RNN-like DeMa is significantly inferior to that of the Transformer-like DeMa, especially in Atari games. These findings suggest that the recurrent mode may be unnecessary in trajectory optimization methods. Given that the hyper-parameters are identical for both types of DeMa except for the sequence length, we assume that variations in sequence length are likely the primary cause of the observed disparities in results. Therefore, we explore the effect of sequence length on the Transformer-like DeMa in subsequent sections.

\begin{table}[htbp]
\vspace{-0.3cm}
  \centering
  \caption{\small The average result of DT, RNN-like DeMa and Transformer-like DeMa in Atari~\cite{atari} and MuJoCo~\cite{todorov2012mujoco}. The results are reported with the normalization following~\cite{ye2021mastering, fujimoto2019off}. Detailed results can be seen in Appendix~\ref{app:comparison}.}
  \small 
    \begin{tabular}{cccc}
    \toprule
    \textbf{Env} & \textbf{DT} & \textbf{RNN-like DeMa} & \textbf{Transformer-like DeMa} \\
    \midrule
    Atari & 62.2  & 67.3  & \textbf{111.8} \\
    MuJoCo & 63.4  & 61.1 & \textbf{66.0} \\
    \bottomrule
    \end{tabular}%
  \label{tab:average}%
 \vspace{-0.2cm}
\end{table}%

\begin{figure}[ht]
    \centering
    \begin{subfigure}[b]{0.32\textwidth}
    \centering
    \includegraphics[width=\linewidth]{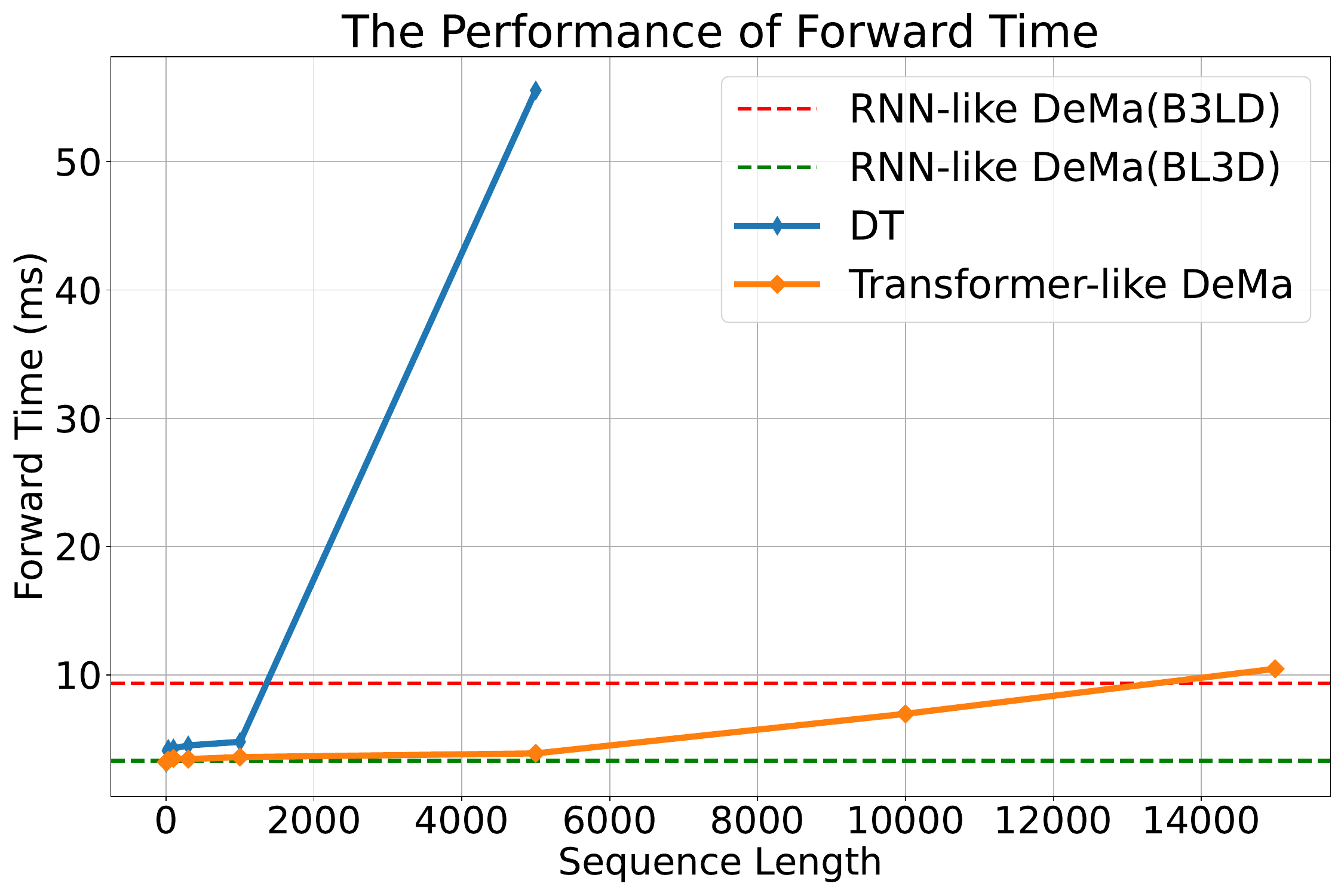}
    \end{subfigure}
    \hfill
    \begin{subfigure}[b]{0.32\textwidth}
    \centering
    \includegraphics[width=\linewidth]{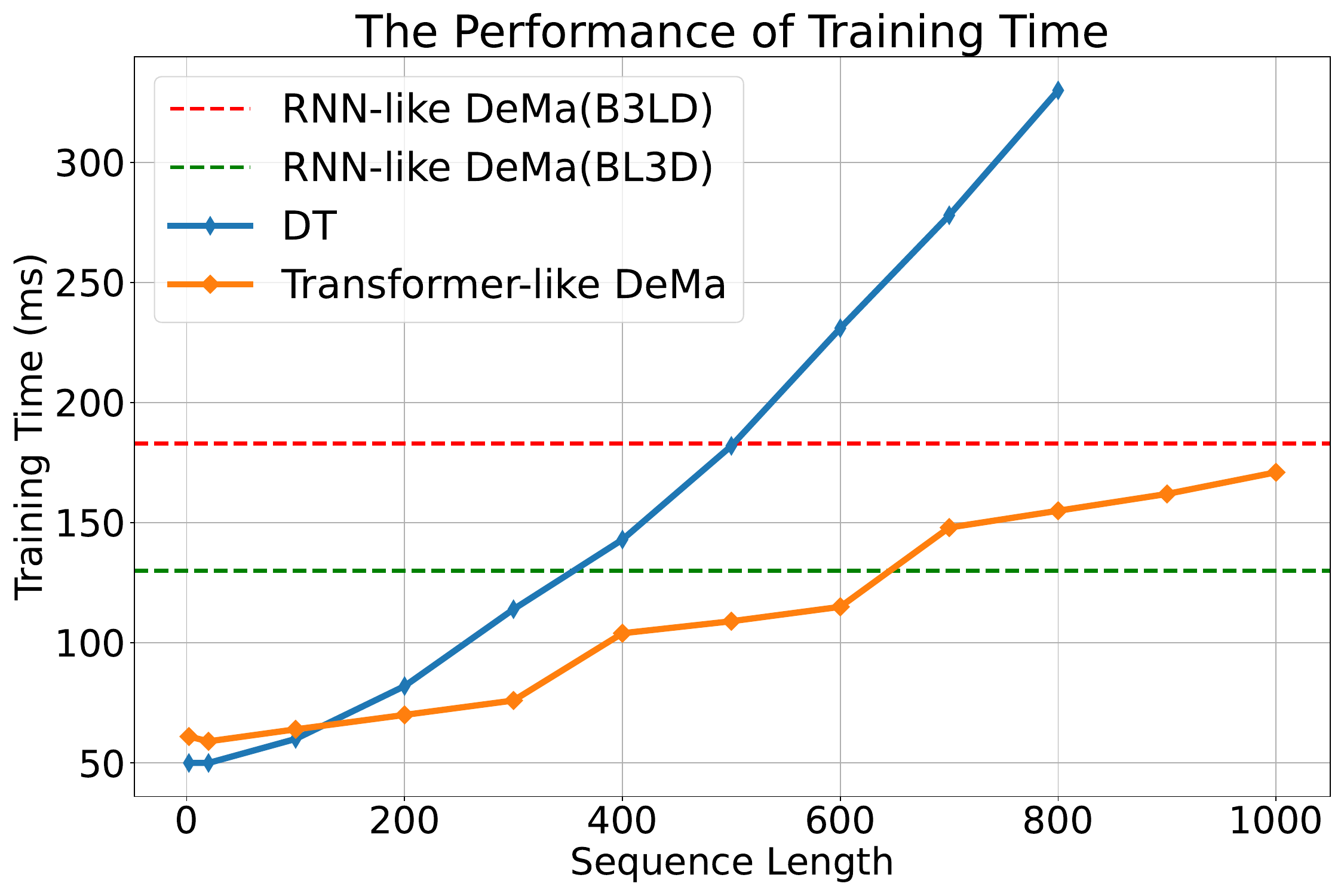}
    \end{subfigure}
    \hfill
    \begin{subfigure}[b]{0.32\textwidth}
    \centering
    \includegraphics[width=\linewidth]{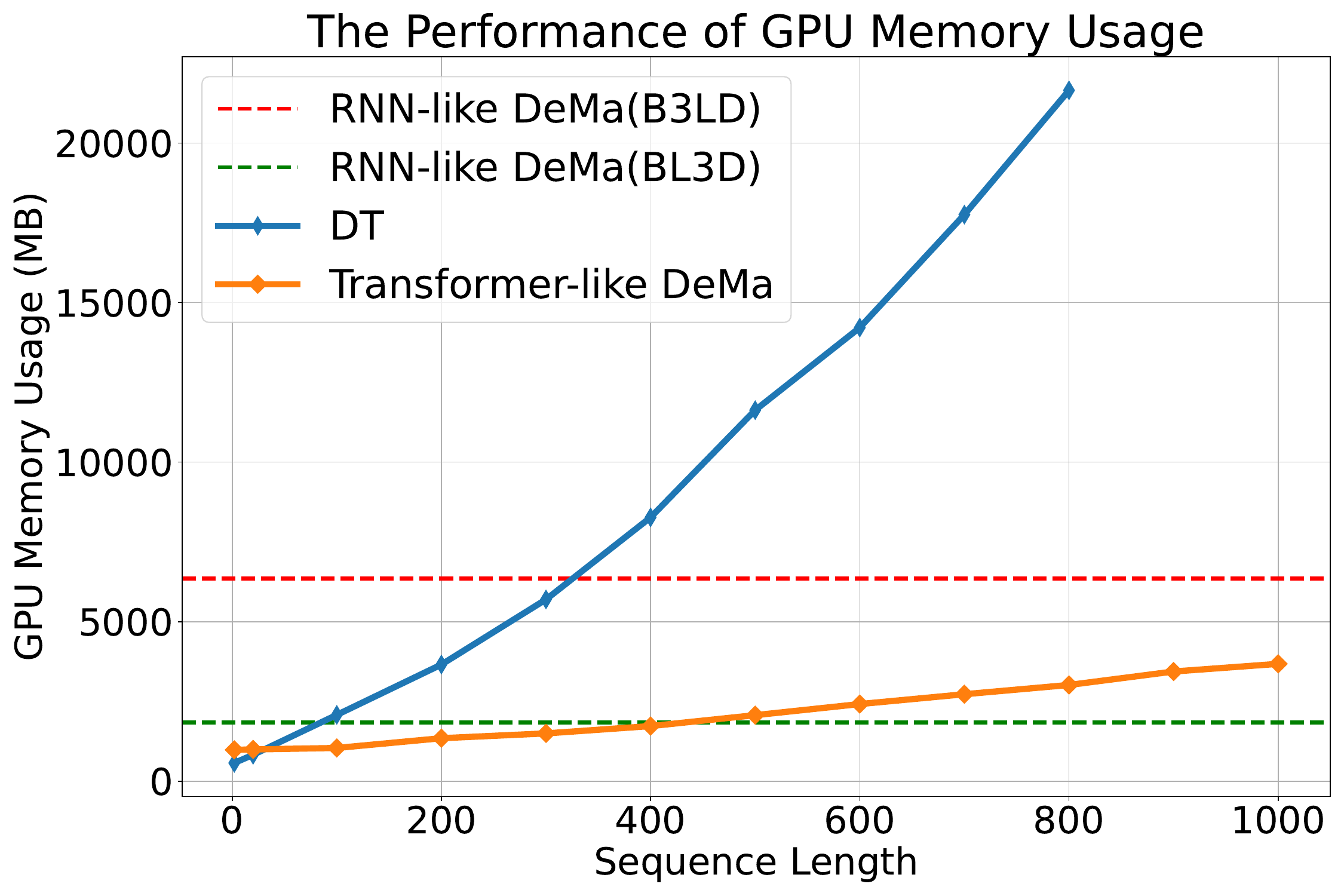}
    \end{subfigure}
    \caption{ \small The impact of sequence length on single-step forward computation time, single-step training time, and GPU memory usage. The sequence length of RNN-like DeMa is 1000.}
    \label{fig:K_seed}
    \vspace{-0.4cm}
\end{figure}

\vspace{-1mm}\paragraph{How does sequence length affect the computational load?}
We investigate the impact of sequence length on single-step training time, single-step inference time and GPU memory usage for models including DT, Transformer-like DeMa, and RNN-like DeMa. Figure~\ref{fig:K_seed} shows that the Transformer-like DeMa operates faster than the RNN-like DeMa when dealing with short sequence lengths, despite that the inference time of RNN-like DeMa is independent of the sequence length. With conventional sequence lengths (such as 20), Transformer-like DeMa holds an advantage in forward speed, training speed, and GPU memory consumption.

\hypothesis{Finding 1}{Transformer-like DeMa is not only faster but also more memory-efficient than RNN-like DeMa for short sequence length. The latter only becomes competitive when processing exceptionally long sequences.} 

\begin{figure}[h]
\vspace{-0.4cm}
    \centering
    \begin{subfigure}[b]{0.4\textwidth}
    \centering
    \includegraphics[width=\linewidth]{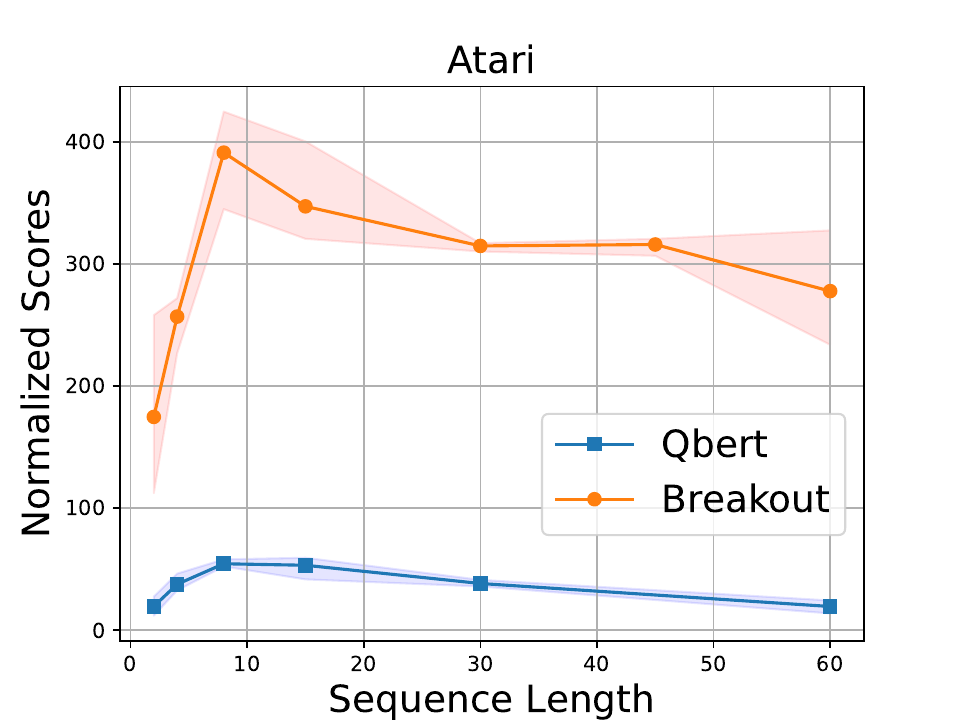}
    \end{subfigure}
    \begin{subfigure}[b]{0.4\textwidth}
    \centering
    \includegraphics[width=\linewidth]{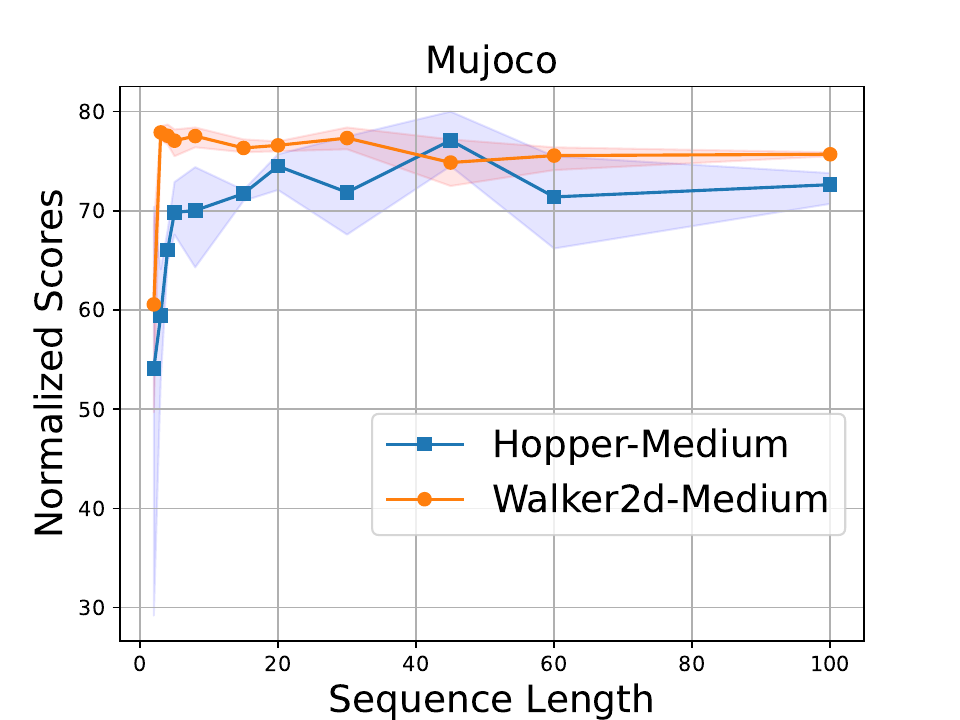}
    \end{subfigure}
    \caption{\small  Comparison of Transformer-like DeMa's Performance on Atari and MuJoCo Tasks. We report mean values averaged over 3 seeds, shaded areas represent deviations.}
    \label{fig:K_performance}
    \vspace{-0.2cm}
\end{figure}

\vspace{-1mm}\paragraph{How does sequence length affect the performance of DeMa?}
While the computational cost of Transformer-like DeMa increases linearly with the expansion of the sequence length, it is crucial to recognize that the increased computational cost may not ensure a corresponding enhancement in the model's performance. Transformer-like DeMa's Performance may plateau or even decline as the input sequence length exceeds a certain threshold. As illustrated in Figure~\ref{fig:K_performance}, Transformer-like DeMa's performance reaches a plateau in MuJoCo~\cite{1606.01540} when the input sequence surpasses a specific length; while significantly deteriorates with excessively long input sequences in Atari.

\hypothesis{Finding 2}{Transformer-like DeMa performs well with a short sequence length. Extending the sequence length beyond an optimal threshold does not yield further improvements and may adversely affect the model's performance.
} 

\begin{figure}[htbp]
    \centering
    \begin{subfigure}[b]{0.23\textwidth}
    \centering
    \includegraphics[width=\linewidth]{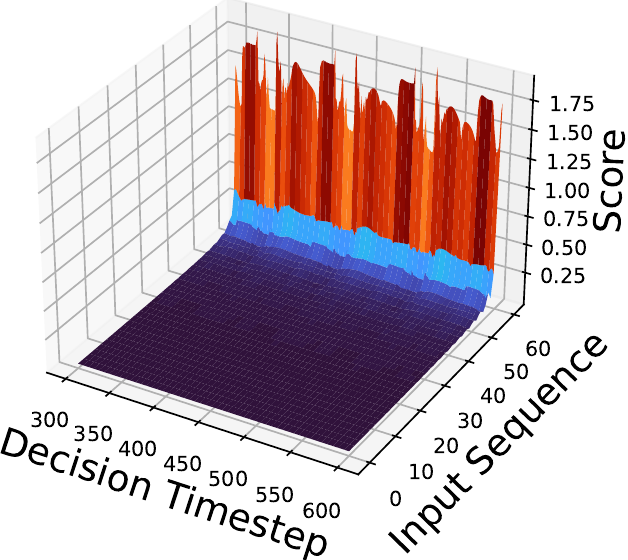}
    \subcaption{Layer 1}
    \end{subfigure}
    \hfill
    \begin{subfigure}[b]{0.23\textwidth}
    \centering
    \includegraphics[width=\linewidth]{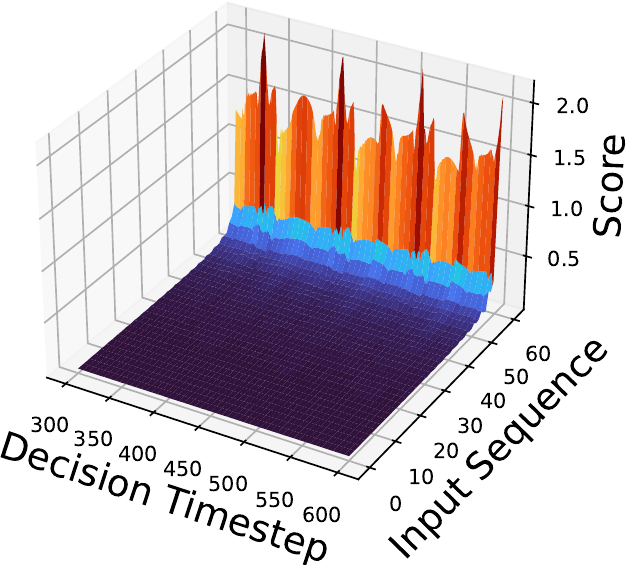}
    \subcaption{Layer 2}
    \end{subfigure}
    \hfill
    \begin{subfigure}[b]{0.23\textwidth}
    \centering
    \includegraphics[width=\linewidth]{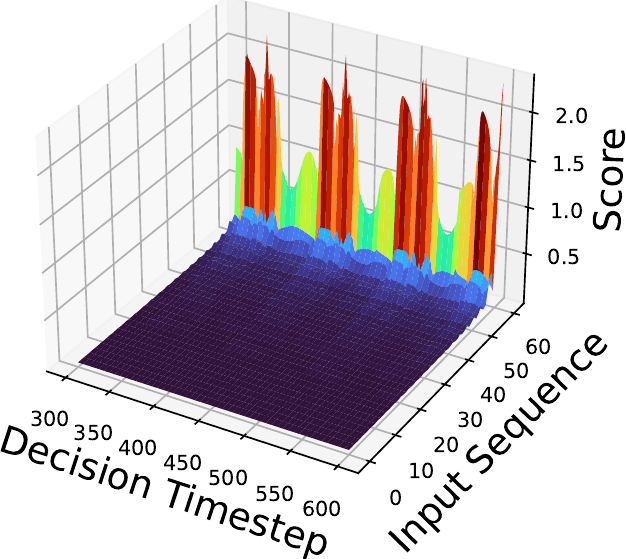}
    \subcaption{Layer 3}
    \end{subfigure}
    \hfill
    \begin{subfigure}[b]{0.23\textwidth}
    \centering
    \includegraphics[width=\linewidth]{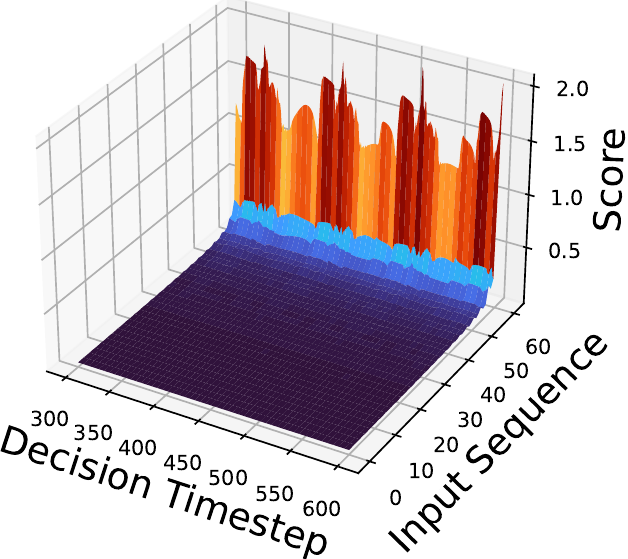}
    \subcaption{Fused Layer}
    \end{subfigure}
    \caption{\small  Hidden attention scores of DeMa from the 300th to the 600th timestep in Hopper-medium-replay. The X-axis represents timesteps from 300 to 600, the Y-axis represents the past $K$ tokens, and the Z-axis indicates the attention scores given to the $K$ tokens at the time of the current decision. More can be seen in Appendix~\ref{app:attention}.}
    \label{fig:attention-mamba}
    \vspace{-0.4cm}
\end{figure}

\vspace{-1mm}\paragraph{Why does DeMa require merely short input sequences?} 
We calculate the hidden attention scores in DeMa via Eq. \eqref{eq:33}-\eqref{eq:44}, which reflect the importance of historical information to DeMa. Figure~\ref{fig:attention-mamba} shows the hidden attention scores of the last $K$ tokens at each decision-making step (from the 300th to the 600th step). It can be seen that the attention scores exhibit exponential decay as the tokens become increasingly distant from the current decision-making moment, which aligns with the forgetting property of a Markov chain~\citep{dc}. What's more, the hidden attention across different decision steps exhibits a periodic pattern towards the current token, suggesting that the model may have learned kinematic features, as agents in these environments engage in periodic movements.\footnote{It is worth noting that what we want to know is the attention scores to the previous $K$ tokens at each decision-making step, which is a bit different from the attention scores between output $y_i$ and input $x_j$, which is explained in detail in Appendix \ref{app:attention}.}

\hypothesis{Finding 3}{The reason that Transformer-like DeMa requires only short input sequences is its hidden attention mechanism primarily focusing on the current token. As a result, Longer sequences can lead to difficulties in training without providing benefits.} 

\begin{wraptable}{r}{0.41\textwidth}
   \vspace{-0.2cm}
    \begin{minipage}{0.41\textwidth}
  \caption{\small Input concatenation types comparison: "BL3D" refers to the concatenation of input tokens across the embedding dimension, while "B3LD" indicates concatenation across the temporal dimension, as depicted in Figure~\ref{fig:struct}. Outcomes are averaged across three random seeds.}
   \label{tab:input tokens}%
  \centering
  \small 
    \begin{tabular}{ccc}
    \toprule
    \textbf{Game} & \textbf{BL3D} & \textbf{B3LD} \\
    \midrule
    Breakout & 72.8±10.6 & \textbf{314.7±10.7} \\
    Qbert & 32.2±14.1 & \textbf{54.4±6.8} \\
    Pong & \textbf{101.9±6.9} & 98.2±12.0 \\
    Seaquest & 1.3±0.0 & \textbf{2.7±0.002} \\
    Asterix & 3.9±0.3 & \textbf{7.8±0.4} \\
    Frostbite & 26.3±20.9 & \textbf{31.1±0.01} \\
    Assault & 127.9±7.1 & \textbf{169.4±33.1} \\
    Gopher & 190.3±60.1 & \textbf{215.8±29.2} \\
    \midrule
    \textbf{Average} & 69.6  & \textbf{111.8 } \\
    \bottomrule
    \end{tabular}
\end{minipage}
\vspace{-1.0cm}
\end{wraptable}
\paragraph{Which type of concatenation is suitable for DeMa?}\label{sec:input type} Models like the Transformer and Mamba typically process inputs token by token. However, given an MDP, there are three elements $s, a, r$ to consider. Therefore a significant design consideration is the method of concatenating these three elements into a suitable token format for the model. We experiment to investigate the suitable design for DeMa. By Table~\ref{tab:input tokens}, concatenating the three elements in the temporal dimension yields better results. This may be due to the significant differences between the three elements of the MDP. As illustrated in~\cite{hu2023graph}, states and actions symbolize fundamentally dissimilar notions, concatenating them in the embedding dimension directly may make it more difficult for the model to recognize, leading to poorer results. 

\hypothesis{Finding 4}{Concatenating state, action, and rtg along the embedding dimension has a significant negative impact on the results.}

\subsection{The Essential Components of DeMa}\label{sec:network}

Aside from the perspective of input data, this section delves into DeMa from the standpoint of network components. We primarily investigate the following questions: (1) Considering that some DTs do not heavily rely on attention mechanism~\cite{dc, lawson2023merging}, is the hidden attention mechanism crucial for DeMa? (2) As the Mamba block is an integration of the hidden attention mechanism with pre up-projection residual blocks~\cite{beck2024xlstm}, what impact will it have on the performance when integrating it with other residual structures (i.e. the post up-projection residual block in the transformer)? (3) With the inherent recurrent nature of SSM~\cite{yu2024mambaout}, does DeMa need position embedding? (Appendix~\ref{app:ablation})

\begin{figure}[h]
\vspace{-0.2cm}
    \centering
\includegraphics[width=1\linewidth]{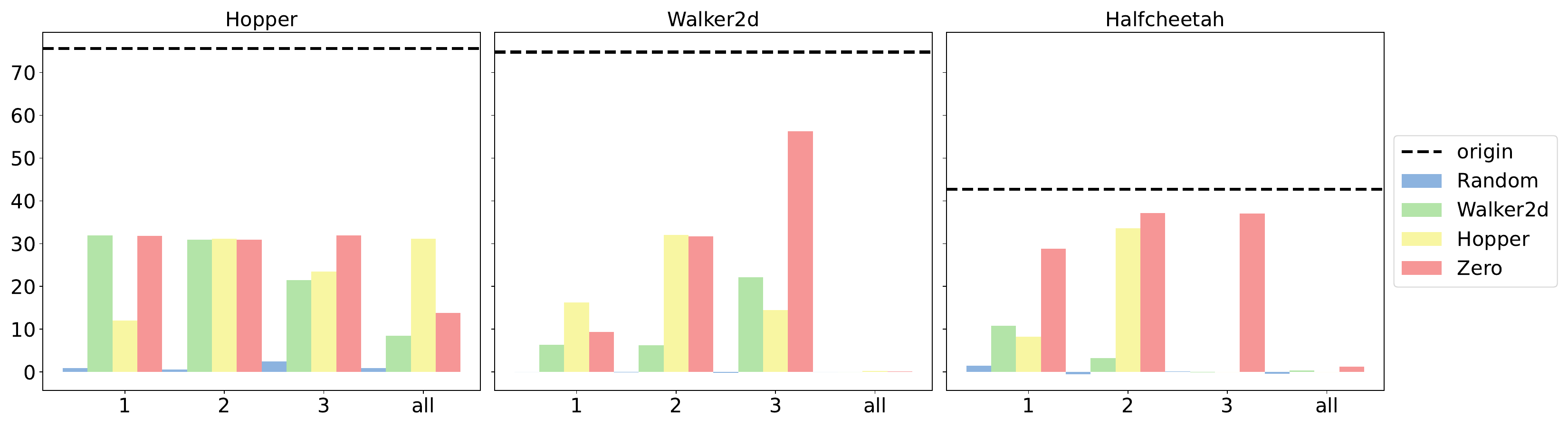}
    \caption{\small Normalized return after swapping the hidden attention of a single layer from another DeMa at a time. The black dashed line represents the evaluation results of the original model. "1", "2", and "3" represent the index of swap layers respectively, and "all" represents the result after swapping all parameters of the hidden attention. It can be seen that swapping the hidden attention has a significant impact on the results.}
    \label{fig:merge}
    \vspace{-0.4cm}
\end{figure}

\paragraph{Is the hidden attention mechanism crucial for DeMa?}
\cite{yu2022metaformer} shows that the transformer does not heavily rely on attention, and~\cite{dc} finds the attention mechanism of DT is not suitable for RL. Given these insights, we aim to investigate whether a similar phenomenon exists in hidden attention. In line with~\cite{lawson2023merging}, we evaluate DeMa by swapping the hidden attention weights trained in different environments, in addition to randomizing and zeroing these weights. As depicted in Figure~\ref{fig:merge}, the performance exhibits a marked decrease regardless of whether the parameters are replaced with those pre-trained in other environments or randomized. Interestingly, when the parameters of hidden attention are set to zero, the model still maintains a certain level of performance. This zeroing of parameters completely removes the hidden attention, ceasing to process historical information and relying solely on residual connections to transmit information. This suggests that the residual connections are functional and the role of hidden attention is crucial for DeMa.

\hypothesis{Finding 5}{Replacing the hidden attention mechanism would lead to a reduction in performance, unlike the attention mechanisms used in transformers. Therefore, the hidden attention mechanism plays a crucial role in DeMa.} 
\vspace{-1mm}\paragraph{What occurs when combining hidden attention with post up-projection residual blocks?}

\begin{table}[h]
\vspace{-0.4cm}
      \caption{\small Performance comparison between DT, hidden attention with post up-projection residual block in the transformer (DeMa with post.) and hidden attention with pre up-projection residual block (DeMa) in Atari. 
      }
      \small
    \label{tab:residual}%
  \centering
    \begin{tabular}{cccc}
    \toprule
    \textbf{Game} & \textbf{DT} & \textbf{DeMa with post.} & \textbf{DeMa} \\
    \midrule
    Breakout & 242.4±31.8 &  296.2±216.3 & \textbf{314.7±10.7} \\
    Qbert & 28.8±10.3 & \textbf{56.9±10.4} & 54.4±6.8 \\
    Pong & \textbf{105.6±2.9} & 104.6±11.9 & 98.2±12.0 \\
    Seaquest & \textbf{2.7±0.7} & 2.6±0.001 & \textbf{2.7±0.002} \\
    Asterix & 5.2±1.2 & 6.5±1.8 & \textbf{7.8±0.4} \\
    Frostbite & 25.6±2.1 & \textbf{31.8±4.8} & 31.1±0.01 \\
    Assault & 52.1±36.2 & 146.4±16.1 & \textbf{169.4±33.1} \\
    Gopher & 34.8±10.0 & \textbf{228.9±81.5} & 215.8±29.2 \\
    \midrule
    \textbf{Average} & 62.2  & 109.2  & \textbf{111.8}  \\
    \bottomrule
    \end{tabular}%
    \caption{\small  Performance comparison between DT, hidden attention with post up-projection residual block in the transformer (DeMa with post.) and hidden attention with pre up-projection residual block (DeMa) in MuJoCo. 
    }
    \label{tab:residual2}%
      \centering
      \small 
    \begin{tabular}{ccccc}
    \toprule
    \textbf{Dataset} & \textbf{Env-Gym} & \textbf{DT} & \textbf{DeMa with post.} & \textbf{DeMa} \\
    \midrule
    M & HalfCheetah & 42.6  & 42.7±0.02 & \textbf{43±0.01} \\
    M & Hopper & 68.4  & 68.4±2.1 & \textbf{74.5±2.9} \\
    M & Walker & 75.5  & \textbf{77.5±2.2} & 76.6±0.2 \\
    \midrule
    M-R & HalfCheetah & 37.0  & \textbf{40.8±0.18} & 40.7±0.03 \\
    M-R & Hopper & 85.6  &  86.1±26.9
 & \textbf{90.7±6.1} \\
    M-R & Walker & 71.2  & \textbf{74.43±2.1} & 70.5±0.1 \\
    \midrule
    M-E & HalfCheetah & 88.8  &  83.8±18 & \textbf{93.2±0.01} \\
    M-E & Hopper & 109.6  & 109.8±0.2 & \textbf{111±0.03} \\
    M-E & Walker & \textbf{109.3 } & \textbf{109.6±0.3} & 106±11.7 \\
    \midrule
    \multicolumn{2}{c}{\textbf{Average-Gym}} & 76.4  & 77.0  & \textbf{78.5 } \\
    \bottomrule
    \end{tabular}%
\end{table}

Mamba represents the integration of the hidden attention mechanism with pre up-projection residual blocks as discussed in~\cite{beck2024xlstm}. To determine the contributing factor to the model's enhanced performance, we explore the combination of hidden attention with post up-projection residual blocks in transformer. According to the results in Table~\ref{tab:residual} and Table~\ref{tab:residual2}, although the overall average results of DeMa are slightly better than those of DeMa with post., it is observable that they each have advantages in different environments. Hence, we believe that the performance differences when integrating with the two types of residual blocks are not statistically significant. It suggests that the structure of the residual blocks exerts minimal influence on the outcome. Given that both configurations yield a measurable performance improvement over the DT, it is reasonable to conclude that the hidden attention mechanism within DeMa plays a pivotal role. 

\hypothesis{Finding 6}{The results obtained using both post up-projection and pre up-projection types of residual block structures are similar while they both perform better than DT. Therefore, the hidden attention mechanism is key to its success.} 

\section{Evaluations on Offline RL Benchmarks}\label{sec:evaluations}

\begin{table}[h]
\vspace{-0.4cm}
  \centering
    \small
  \caption{\small Results for 1\% DQN-replay datasets. We evaluate the performance of DeMa on eight Atari games.}
    \begin{tabular}{ccccccc}
    \toprule
    \textbf{Game} & \textbf{CQL} & \textbf{BC} & \textbf{DT} & \textbf{DC} & \textbf{DC}$^{hybrid}$ & \textbf{DeMa(Ours)} \\
    \midrule
    Breakout & 211.1  & 142.7  & 242.4±31.8 & 352.7±44.7 & \textbf{416.0 ±105.4} & 314.7±10.7 \\
    Qbert & \textbf{104.2 } & 20.3  & 28.8±10.3 & 67.0±14.7 & 62.6 ±9.4 & 54.4±6.8 \\
    Pong & \textbf{111.9 } & 76.9  & 105.6±2.9 & 106.5±2.0 & 111.1 ±1.7 & 98.2±12.0 \\
    Seaquest & 1.7  & 2.2  & \textbf{2.7±0.7} & 2.6±0.3 & \textbf{2.7 ±0.04} & \textbf{2.7±0.002} \\
    Asterix & 4.6  & 4.7  & 5.2±1.2 & 6.5±1.0 & 6.3 ±1.8 & \textbf{7.8±0.4} \\
    Frostbite & 9.4  & 16.1  & 25.6±2.1 & 27.8±3.7 & 28.0 ±1.8 & \textbf{31.1±0.01} \\
    Assault & 73.2  & 62.1  & 52.1±36.2 & 73.8±20.3 & 79.0 ±13.1 & \textbf{169.4±33.1} \\
    Gopher & 2.8  & 33.8  & 34.8±10.0 & 52.5±9.3 & 51.6 ±10.7 & \textbf{215.8±29.2} \\
    \midrule
    \textbf{Average} & 64.9  & 44.9  & 62.2  & 86.2  & 94.7  & \textbf{ 111.8 } \\
    \bottomrule
    \end{tabular}%
  \label{tab:1111}%
  \label{tab:atari result}%

  \centering
    \small
  \caption{\small  Results for MuJoCo. The dataset names are abbreviated as follows: "medium" as "M", "medium-replay" as "M-R" and "medium-expert" as "M-E". The results are reported with the expert-normalized following~\cite{fujimoto2019off}.}
  \small 
    \begin{tabular}{cccccccc}
    \toprule
    \textbf{Dataset} & \textbf{Environment} & \textbf{CQL} & \textbf{DS4} & \textbf{RvS} & \textbf{DT} & \textbf{GDT} & \textbf{DeMa(Ours)} \\
    \midrule
    M & HalfCheetah & 44.0  & 42.5  & 41.6 & 42.6  & 42.9  & 43±0.01 \\
    M & Hopper & 58.5  & 54.2  & 60.2  & 68.4  & 65.8  & \textbf{74.5±2.9} \\
    M & Walker & 72.5  & 78.0  & 71.7  & 75.5  & 77.8  & 76.6±0.2 \\
    \midrule
    M-R & HalfCheetah & 45.5  & 15.2  & 38 & 37.0  & 39.9  & 40.7±0.03 \\
    M-R & Hopper & 95.0  & 49.6  & 73.5 & 85.6  & 81.6  & \textbf{90.7±6.1} \\
    M-R & Walker & 77.2  & 69.0  & 60.6 & 71.2  & \textbf{74.8 } & 70.5±0.1 \\
    \midrule
    M-E & HalfCheetah & 91.6  & 92.7  & 92.2  & 88.8  & 92.4  & \textbf{93.2±0.01} \\
    M-E & Hopper & 105.4  & 110.8  & 101.7  & 109.6  & 110.9  & \textbf{111±0.03} \\
    M-E & Walker & 108.8  & 105.7  & 106.0  & \textbf{109.3 } & \textbf{109.3 } & 106±11.7 \\
    \midrule
    \multicolumn{2}{c}{\textbf{Average}} & 77.6  & 68.6  & 71.7  & 76.4  & 76.8  & \textbf{78.5 } \\
    \bottomrule
    \end{tabular}%
  \label{tab:MuJoCo}%

  \centering
    \small
  \caption{\small  The resource usage for training DT,
DC and DeMa on Atari and MuJoCo.}
\small 
    \begin{tabular}{ccccc}
    \toprule
      & \textbf{Complexity} & \textbf{DT} & \textbf{DC} & \textbf{DeMa(Ours)} \\
    \midrule
      & Training time per step(ms) & 55 & \textbf{43} & 50 \\
    \textbf{Atari} & GPU memory usage(GiB) & 4.2 & \textbf{3.0}  & 4.2 \\
      & MACs & 12.1G/46.5G  & 11.1G/40.6G  & \textbf{8.8G/36.3G} \\ 
      & All params \# & 2.35M & 1.94M & \textbf{1.7M} \\
    \midrule
      & Training time per step(ms) & 56/58 & \textbf{53.6/53.9} & 57.6/58.8 \\
    \textbf{Gym} & GPU memory usage(GiB) & 0.65/0.8 & \textbf{0.55/0.6} & 1.0/1.0 \\
      & MACs & 2.5G/9.5G & 1.6G/6.1G & \textbf{0.7G/2.1G} \\
      & All params \# & 726.2K/2.6M & 536K/1.9M & \textbf{175.5K/500.0K} \\
    \bottomrule
    \end{tabular}%
  \label{tab:param}%
\end{table}%

In this section, we delve into a comparative analysis of DeMa's performance against various DTs.  Our investigation primarily centers on the influence of disparate network architectures on the experimental outcomes. Consistent with antecedent studies, we assessed both discrete (Atari~\cite{atari}) and continuous control tasks (MuJoCo~\cite{d4rl}), presenting the normalized scores accordingly. Given that the sequence length considerably affects the results, we selected the optimal outcomes from sequence lengths $K=8$ to $K=20$ for DeMa. The detailed hyper-parameters on DeMa are available in Appendix~\ref{app:hyperparameter}. Our main results are shown in Table~\ref{tab:atari result} and Table~\ref{tab:MuJoCo}. DeMa achieves a significantly higher average score compared to DT in Atari games, while the number of parameters and the number of MACs in DeMa are each five times fewer than those in DT, as shown in Table~\ref{tab:param}. Moreover, DeMa has better scalability for input length which can be seen in Figure~\ref{fig:K_seed}, it maintains a slow linear growth with the input sequence length increases while the computational cost of the Transformer grows quadratically. These results demonstrate that our transformer-like DeMa is well-suited for integration with trajectory optimization methods.

\section{Conclusion}\label{sec:conclusion}

To investigate Mamba's compatibility with trajectory optimization, this work conducts comprehensive experiments from the aspect of data structures and network architectures. Our findings reveal that (1) DeMa benefits from short sequence lengths due to its exponentially decaying focus on sequences. Consequently, we incorporate a Transformer-like DeMa. (2) The hidden attention mechanism plays a crucial role in DeMa. It can combine with other residual structures and does not require position embedding. Based on the insights gained from the investigation, our DeMa surpasses previous methods, achieving higher performance over the DT while using 30\% fewer parameters in eight Atari games. In the MuJoCo, our DeMa outperforms DT with only a quarter of the parameters. In conclusion, our DeMa is compatible with trajectory optimization in offline RL.

\textbf{Limitations.}  We investigate the application of Mamba in trajectory optimization and present findings that provide valuable insights for the community. However, there remain several limitations: (1) Trajectory optimization tasks typically involve shorter input sequences, raising questions about how well the RNN-like DeMa performs in terms of memory capacity in RL compared to models such as RNNs and LSTMs. Furthermore, the potential of both types of DeMa warrants further exploration, particularly in some POMDP environments and long-horizon non-Markovian tasks that require long-term decision-making and memory. (2) We examine the importance of the hidden attention mechanism in Section \ref{sec:network}, future work could leverage interpretability tools to examine further the causal relationship between memory and current decisions in DeMa, ultimately contributing to the development of interpretable decision models. (3) While we have assessed the properties of DeMa and identified improvements in both performance efficiency and model compactness compared to DT, it remains unclear whether DeMa is suitable for multi-task RL and online RL environments.

\begin{ack}
This work is supported by STI 2030--Major Projects (No. 2021ZD0201405), National Natural Science Foundation of China (No. 72301289), and the Zhejiang Province Science Foundation under Grants LD24F020002. We thank zigzagcai for his \href{https://github.com/state-spaces/mamba/pull/244}{PR}: support variable-length sequences for mamba block.  We thank Liang Zhang and Yang Ma for their valuable suggestions and collaboration. We sincerely appreciate the time and effort invested by the anonymous reviewers in evaluating our work and are grateful for their valuable and insightful feedback.
\end{ack}


{
\small
\bibliographystyle{unsrt}
\bibliography{references}
}


\clearpage

\appendix
\vspace{7pt}
\section*{\Large \centering Supplementary Material for  \\ 
\vspace{7pt}
{Is Mama Compatible with Trajectory Optimization in Offline Reinforcement Learning?}}

\vspace{20pt}

\addtocontents{toc}{\setcounter{tocdepth}{1}} 

\tableofcontents

\section{Environment and Dataset}

\paragraph{MuJoCo.} The MuJoCo domain~\cite{todorov2012mujoco} evaluates the performance of RL algorithms in continuous control tasks. In keeping with previous studies, we select three games from the standard locomotion environments~\cite{todorov2012mujoco} in Gym~\cite{1606.01540}, namely HalfCheetah, Hopper, and Walker, and three different dataset settings, namely
medium, medium-replay, and medium-expert~\cite{d4rl}.

\paragraph{Atari.} Atari~\cite{atari} is an ideal platform for evaluating an agent's ability in long-term credit assignments. We conduct experiments in eight different games: Breakout, Qbert, Pong, Seaquest, Asterix, Frostbite, Assault, and Gopher. We use 1\% DQN Replay Dataset~\cite{agarwal2020optimistic} as our training dataset, which encompasses a total of 500,000 timesteps worth of samples generated throughout the training process of a DQN agent~\cite{mnih2015human}. It's worth noting that the version of "atari-py" and "gym" we use is 0.2.5 and 0.19.0 respectively, which is noted by the official code in \href{https://github.com/google-research/batch\_rl}{https://github.com/google-research/batch\_rl}.

\section{Baselines}

\paragraph{Baselines for MuJoCo.} To evaluate DeMa’s performance in the MuJoCo, we compare DeMa with one value-based method: CQL~\cite{kumar2020conservative} and four trajectory optimization methods with different network architectures: DS4~\cite{david2022ds4}, RvS~\cite{afsar2022reinforcement}, DT~\cite{chen2021decisiontransformer}, GDT~\cite{hu2023graph} and obtain baseline performance scores for CQL and DS4 from~\cite{dc}, for RvS from~\cite{afsar2022reinforcement} and for GDT from~\cite{hu2023graph}.

\paragraph{Baselines for Atari.} In the Atari domain, we compare DeMa with CQL~\cite{kumar2020conservative}, DT~\cite{chen2021decisiontransformer}, DC and DC$^{hybrid}$~\cite{dc}. The results of baselines are directly borrowed from~\cite{dc}.

\section{The Procedure of Training and Inference}\label{app:procedure}

\paragraph{Training resources}

We use one NVIDIA GeForce RTX 4090 to train each model in MuJoCo and one NVIDIA GeForce RTX 3090 to train each model in Atari. Training each model typically takes 3-8 hours and 5-14 hours in MuJoCo and Atari respectively. However, since each environment needs to be trained three
times with different seeds, the total training time is usually multiplied by three.

\paragraph{The procedure of Transformer-like DeMa}
The Training and evaluation for Transformer-like DeMa are similar to variant DTs. Given a dataset of offline trajectories, we randomly select a starting point and truncate it into a sequence of length $K$. After forming a batch of data, it is input into the model for training. We minimize the reconstruction loss between the predicted action and the actual action, i.e. the cross-entropy loss for discrete actions and Mean Square Error (MSE) for continuous actions. The input data is also a sequence of length $K$ in the evaluation phase.

\paragraph{The procedure of RNN-like DeMa}
For RNN-like DeMa, the input during training is a batch of complete trajectories. As different trajectories have different lengths, we pad the trajectories to the same length before inputting them into the model and mask the loss of the padding. However, training with full trajectories rather than truncated sequences may be more inefficient, especially in scenarios where sequence lengths vary widely. In the DQN Replay Dataset in Atari, the lengths of different trajectories varied dramatically. Some trajectories might only be 500 timesteps long, while others could contain a sample with a length of 10,000 timesteps. This causes a lot of computing resources to be wasted on meaningless padding, resulting in inefficiency and ineffectiveness. Some techniques can avoid this issue. One can refer to this \href{https://github.com/state-spaces/mamba/pull/244}{PR}.

\section{Implementation details of DeMa}\label{app:hyperparameter}

We implement DeMa based on the official code of \href{https://github.com/kzl/decision-transformer}{DT} and the \href{https://github.com/state-spaces/mamba.git}{Mamba}. We have also adopted the code from \href{https://github.com/AmeenAli/HiddenMambaAttn?tab=readme-ov-file}{HiddenMambaAttn} to calculate the attention scores of DeMa on the current input sequence at each decision step. Given that the official Mamba code utilizes Triton, we also employ \href{https://github.com/johnma2006/mamba-minimal/tree/master}{Mamba-minimal} which is fully based on pytroch to compute the MACs of DeMa. 

Tables~\ref{tab:hyper1}-\ref{tab:hyper3} provide a comprehensive list of hyper-parameters for our proposed transformer-like DeMa and RNN-like DeMa applied to MuJoCo and Atari environments. To
ensure a fair comparison, we adopt similar hyper-parameter settings to DT~\cite{chen2021decisiontransformer} and DC~\cite{dc}.

\subsection{Hyper-parameters in MuJoCo}

\begin{table}[h]
  \centering
  \caption{Hyper-parameters of DeMa for MuJoCo.}
    \begin{tabular}{cl}
    \toprule
    Hyper-parameter & \multicolumn{1}{c}{Value} \\
    \midrule
    Layers & \multicolumn{1}{c}{3} \\
    Embedding dimension & \multicolumn{1}{c}{128} \\
    Nonlinearity function & \multicolumn{1}{c}{\textbackslash{}} \\
    Batch size & \multicolumn{1}{c}{64} \\
    Context length $K$ & \multicolumn{1}{c}{20} \\
    Dropout & \multicolumn{1}{c}{0.0} \\
    Learning rate & \multicolumn{1}{c}{$10^{-4}$} \\
    Grad norm clip & \multicolumn{1}{c}{0.25} \\
    Weight decay & \multicolumn{1}{c}{$10^{-4}$} \\
    Learning rate decay & Linear warmup for first $10^5$ training steps \\
    d\_model & \multicolumn{1}{c}{64} \\
    d\_state & \multicolumn{1}{c}{64} \\
    expand & \multicolumn{1}{c}{2} \\
    \bottomrule
    \end{tabular}%
  \label{tab:hyper1}%
\end{table}%

For our training on MuJoCo, the majority of the hyper-parameters in Table~\ref{tab:hyper1} are adapted from~\cite{dc}. For the learning rate, we use a learning rate of $10^{-4}$ for training in hopper-medium, hopper-medium-replay, and walker2d-medium and use $10^{-3}$ for other environments. For the embedding dimension, we use an embedding dimension of 256 in hopper-medium and hopper-medium-replay, while use 128 in the other environments. What's more, as DeMa does not use multilayer perceptron (MLP), so there is no nonlinearity function for DeMa. As for DeMa with post. in Table~\ref{tab:residual}, we use ReLU as per convention. For DeMa's hyper-parameters, we use a d\_model of 128 in all expert datasets, while use 64 in the other environments. As for d\_state and expand, we set 64 and 2 respectively for all env. We keep the experimental parameters consistent for all types of DeMa in MuJoCo.

\subsection{Hyper-parameters in Atari}

\paragraph{Transformer-like DeMa.} For the Atari game we mostly follow those in Table~\ref{tab:hyper2} from~\cite{chen2021decisiontransformer}. The only adjustment made is to the context length $K$ and return-to-go conditioning. As revealed in Figure~\ref{fig:K_performance}, the sequence length is not always better when it's longer. Thus for Qbert and Frostbite we use $K=8$. For other games, we keep $K=30$. As for the return-to-go conditioning, we find the return obtained by DeMa in some games has already exceeded the initial "return to go" set for DT. Therefore, we increase the "return to go" so that DeMa can fully demonstrate its performance.
 
\begin{table}[h]
  \centering
  \caption{Hyper-parameters of Transformer-like DeMa for Atari.}
    \begin{tabular}{cc}
    \toprule
    Hyper-parameter & Value \\
    \midrule
    Layers & 6 \\
    Embedding dimension & 256 \\
    Nonlinearity function & ReLU(state encoder) \\
    Batch size & 128 \\
    Context length $K$ & 30 \\
    Return-to-go conditioning & 90 Breakout,12000 Qbert\\
    & 20 Pong,1750 Seaquest \\
      & 700 Asterix, 1450 Frostbite\\
      &1200 Assault, 6500 Gopher \\
    Dropout & 0.1 \\
    Learning rate & $6\times 10^{-4}$ \\
    Grad norm clip & 1 \\
    Weight decay & 0.1 \\
    Learning rate decay & Linear warmup and cosine decay (see code for details) \\
    Max epochs & 10 \\
    Adam betas & (0.9, 0.95) \\
    Warmup tokens & 512$\times$ 20 \\
    Final tokens & 6$\times$ 500000$\times$ K \\
    d\_model & 128 \\
    d\_conv & 4 \\
    d\_state & 64 \\
    expand & 2 \\
    \bottomrule
    \end{tabular}%
  \label{tab:hyper2}%
\end{table}%

\begin{table}[h]
  \centering
  \caption{Hyper-parameters of RNN-like DeMa for Atari. The other hyper-parameters are kept consistent with those in Table~\ref{tab:hyper2}.}
    \begin{tabular}{cc}
    \toprule
    Hyper-parameter & Value \\
    \midrule
    Context length & all trajectory \\
    Batch size & 8 \\
    Learning rate & $10^{-4}$ \\
    inner\_it & 200 \\
    \bottomrule
    \end{tabular}%
  \label{tab:hyper3}%
\end{table}%

\paragraph{RNN-like DeMa.} Since the RNN-like DeMa utilizes trajectories for training and the trajectories in Atari are exceptionally lengthy, the available sample size becomes significantly limited when only 1\% of the DQN-replay dataset is utilized. If the prior parameter settings were to be used, the training would done after only a few hundred upgrades, thereby resulting in an unsatisfactory performance. Therefore, we consider multiple updates for a single sample, while simultaneously lowering the learning rate as shown in Table~\ref{tab:hyper3}. What's more, due to the limitation of GPU memory, we can only set a batch size of 8 for Atari. Specifically, For the Frostbite, we set a batch size of 1, an epoch of 50. The other hyper-parameters are kept consistent with those in Table~\ref{tab:hyper2}.

\section{Detailed results}\label{app:comparison}

Table~\ref{tab:RNN1} and Table~\ref{tab:compare two mamba} show detailed results between RNN-like DeMa\footnote{Due to the high experimental costs, we only run the RNN-like DeMa in Atari once.} and Transformer-like DeMa. It can be observed that the performance of the RNN-like DeMa is not as good as that of the Transformer-like DeMa, and Figure~\ref{fig:K_seed} also shows that the RNN-like DeMa requires more computational overhead. Hence, using the RNN model in trajectory optimization seems to be unnecessary, as section~\ref{sec:RNN-like} finds that past historical information does not provide much assistance to current decision-making. However, in tasks that require memory capability or are model-based, the RNN-like DeMa could be a better choice. This could be a direction for deeper future research based on~\cite{samsami2024mastering, deng2024facing, lu2024structured}.

\begin{table}[h]
  \centering
  \small
  \caption{The Comparison of DT, RNN-like DeMa, and Transformer-like DeMa in Atari Games.}
    \begin{tabular}{cccc}
    \toprule
    \textbf{Env} & \textbf{DT} & \textbf{RNN-like DeMa} & \textbf{Transformer-like DeMa} \\
    \midrule
    Breakout & 242.4±31.8 & 166.0  & \textbf{314.7±10.7} \\
    Qbert & 28.8±10.3 & 13.6  & \textbf{54.4±6.8} \\
    Pong & 105.6±2.9 & \textbf{109.6 } & 98.2±12.0 \\
    Seaquest & \textbf{2.7±0.7} & 1.7  & \textbf{2.7±0.002} \\
    Asterix & 5.2±1.2 & 4.7  & \textbf{7.8±0.4} \\
    Frostbite & 25.6±2.1 & 8.6  & \textbf{31.1±0.01} \\
    Assault & 52.1±36.2 & 117.8  & \textbf{169.4±33.1} \\
    Gopher & 34.8±10.0 & 116.7  & \textbf{215.8±29.2} \\
    \midrule
    \textbf{Average} & 62.2  & 67.3  & \textbf{111.8} \\
    \bottomrule
    \end{tabular}%
  \label{tab:RNN1}%
      \vspace{-0.4cm}
\end{table}%

\begin{table}[htp]
  \centering
    \small
  \caption{The comparison between DT, RNN-like DeMa, and Transformer-like DeMa in MuJoCo.}
    \begin{tabular}{ccccc}
    \toprule
    \textbf{Dataset} & \textbf{Environment} & \textbf{DT} & \textbf{RNN-like DeMa} & \textbf{Transformer-like DeMa} \\
    \midrule
    M & HalfCheetah & 42.6  & 42.6±0 & \textbf{43±0.01} \\
    M & Hopper & 68.4  & 61.7±4.9 & \textbf{74.5±2.9} \\
    M & Walker & 75.5  & \textbf{76.7±0.2} & 76.6±0.2 \\
    \midrule
    M-R & HalfCheetah & 37.0  & 36.9±0.3 & \textbf{40.7±0.03} \\
    M-R & Hopper & 85.6  & 80.5±25.8 & \textbf{90.7±6.1} \\
    M-R & Walker & \textbf{71.2 } & 68.1±8.1 & \textbf{70.5±0.1} \\
    \midrule
    \multicolumn{2}{c}{\textbf{Average}} & 63.4  & 61.1 & \textbf{66.0}  \\
    \bottomrule
    \end{tabular}%
  \label{tab:compare two mamba}%
      \vspace{-0.4cm}
\end{table}%

\section{Tasks Requires Long Horizon Planning Skills}

Three experiments involving delayed rewards(MuJoCo with delayed rewards) and maze navigation(maze2d, antmaze) are conducted to investigate how would DeMa perform on tasks that require long horizon planning skills.

\subsection{MuJoCo with Delayed Rewards}

\begin{table}[h]
  \centering
  \caption{Results for D4RL datasets with delayed (sparse) reward. The "Origin Average" in the table represents the normalized scores of evaluations across six datasets under the original dense reward setting. }
    \begin{tabular}{ccccccc}
    \toprule
    \textbf{Dataset} & \textbf{Env-Gym} & \textbf{CQL} & \textbf{DS4} & \textbf{DT} & \textbf{GDT} & \textbf{DeMa(Ours)} \\
    \midrule
    M & HalfCheetah & 1.0 ± 1.0 & 42.7±0 & 42.2 ± 0.2 & \textbf{43±0} & 42.9±0.01 \\
    M & Hopper & 23.3 ± 1.0 & 58.2±0.7 & 57.3 ± 2.4 & 58.2±2.4 & \textbf{69.1±6.5} \\
    M & Walker & 0.0 ± 0.4 & 75.7±0.5 & 69.9 ± 2.0 & \textbf{78.9±0.1} & 77.6±1.5 \\
    \midrule
    M-R & HalfCheetah & 7.8 ± 6.9 & 15.5±0 & 33.0 ± 4.8 & 41±0.1 & \textbf{41.1±0.15} \\
    M-R & Hopper & 7.7 ± 5.9 & 77.5±0.4 & 50.8 ± 14.3 & 79.8±15.9 & \textbf{83.8±6.9} \\
    M-R & Walker & 3.2 ± 1.7 & 69.1±3.2 & 51.6 ± 24.6 & 70.4±8.7 & \textbf{71.7±5} \\
    \midrule
    \multicolumn{2}{c}{\textbf{Average-Gym}} & 7.2 & 56.45 & 50.8 & 61.9  & 64.4  \\
    \bottomrule
    \end{tabular}%
  \label{tab:delayed}%
\end{table}%

To investigate DeMa's performance on tasks with delayed rewards, we conduct an experiment on a delayed return version of the D4RL benchmarks~\cite{chen2021decisiontransformer}, in which the agent does not receive any rewards along the trajectory but instead receives the cumulative reward of the trajectory in the final timestep. In this environment, we train DeMa using the same hyper-parameters settings, and the results are shown in Table \ref{tab:delayed}. Results show that CQL is the most affected, while DT also experiences a certain degree of influence. In contrast, DeMa is relatively less impacted. The results indicate that DeMa demonstrates effective performance in tasks with delayed rewards.

\subsection{Maze Navigation}

There are two environments in maze navigation. \textbf{Maze2d}: This environment aims at reaching goals with sparse rewards, which is suitable for assessing the model’s capability to efficiently integrate data and execute long-range planning. The objective of this domain is to guide an agent through a maze to reach a designated goal. \textbf{Antmaze}: This environment is similar to maze2d, while the agent is an ant with 8 degrees of freedom. For our training on Maze, the majority of the hyper-parameters in Table~\ref{tab:hyper antmaze} and Table~\ref{tab:hyper maze2d}. For maze2d-medium, we use $K=8$ and embedding\_dim=256. For maze2d-umaze, we use hyper-parameters in Table~\ref{tab:hyper1}.

\begin{table}[h]
\vspace{-0.5cm}
  \centering
  \caption{Hyper-parameters of DeMa for antmaze.}
    \begin{tabular}{cl}
    \toprule
    Hyper-parameter & \multicolumn{1}{c}{Value} \\
    \midrule
    Layers & \multicolumn{1}{c}{3} \\
    Embedding dimension & \multicolumn{1}{c}{128} \\
    Nonlinearity function & \multicolumn{1}{c}{\textbackslash{}} \\
    Batch size & \multicolumn{1}{c}{32} \\
    Context length $K$ & \multicolumn{1}{c}{5} \\
    Dropout & \multicolumn{1}{c}{0.1} \\
    Learning rate & \multicolumn{1}{c}{$2e^{-5}$} \\
    Grad norm clip & \multicolumn{1}{c}{0.25} \\
    Weight decay & \multicolumn{1}{c}{$10^{-4}$} \\
    Learning rate decay & Linear warmup for first $10^5$ training steps \\
    d\_model & \multicolumn{1}{c}{128} \\
    d\_state & \multicolumn{1}{c}{64} \\
    num\_eval\_episodes & \multicolumn{1}{c}{50} \\
    max\_iters & \multicolumn{1}{c}{50} \\
    num\_steps\_per\_iter & \multicolumn{1}{c}{2000} \\
    \bottomrule
    \end{tabular}%
  \label{tab:hyper antmaze}%
  \caption{Hyper-parameters of DeMa for maze2d.}
    \begin{tabular}{cl}
    \toprule
    Hyper-parameter & \multicolumn{1}{c}{Value} \\
    \midrule
    Layers & \multicolumn{1}{c}{3} \\
    Embedding dimension & \multicolumn{1}{c}{128} \\
    Nonlinearity function & \multicolumn{1}{c}{\textbackslash{}} \\
    Batch size & \multicolumn{1}{c}{32} \\
    Context length $K$ & \multicolumn{1}{c}{20} \\
    Dropout & \multicolumn{1}{c}{0.1} \\
    Learning rate & \multicolumn{1}{c}{$2e^{-5}$} \\
    Grad norm clip & \multicolumn{1}{c}{0.25} \\
    Weight decay & \multicolumn{1}{c}{$10^{-4}$} \\
    Learning rate decay & Linear warmup for first $10^5$ training steps \\
    d\_model & \multicolumn{1}{c}{64} \\
    d\_state & \multicolumn{1}{c}{64} \\
    num\_eval\_episodes & \multicolumn{1}{c}{50} \\
    max\_iters & \multicolumn{1}{c}{50} \\
    num\_steps\_per\_iter & \multicolumn{1}{c}{2000} \\
    \bottomrule
    \end{tabular}%
  \label{tab:hyper maze2d}%
\end{table}%

We compare DeMa with DT~\cite{chen2021decisiontransformer}, GDT~\cite{hu2023graph} and DC~\cite{dc}. The results of DT and GDT are directly borrowed from~\cite{QT}. Results in Table \ref{tab:maze} show that DeMa performs better compared to DT in the maze navigation task. The visualization analysis of the hidden attention mechanism in these environments can be found in Figure~\ref{fig:attention-mamba5} and Figure~\ref{fig:attention-mamba6}.

\begin{table}[htbp]
  \centering
  \caption{Results for maze2d and antmaze.}
    \begin{tabular}{cccccc}
    \toprule
    \textbf{Dataset} & \textbf{Env-Gym} & \textbf{DT} & \textbf{GDT} & \textbf{DC} & \textbf{DeMa(Ours)} \\
    \midrule
    umaze &  & 31.0 & 50.4 & 36.3±3 & 54.3±9.4 \\
    medium & maze2d & 8.2 & 7.8 & 2.1±1.02 & 10.3±3.1 \\
    large &   & 2.3 & 0.7 & 0.9±0 & 2.8±2.2 \\
    \midrule
    umaze & {antmaze} & 59.2 & 76 & 85.00 & 82±0 \\
    umaze-diverse &   & 53 & 69 & 78.5  & 80.7±6.2 \\
    \bottomrule
    \end{tabular}%
  \label{tab:maze}%
\end{table}%

\section{Further Ablation Study}\label{app:ablation}
\vspace{-1mm}\paragraph{DeMa does not need the position embedding.}

\begin{table}[h]
  \centering
    \small
  \caption{The affection of position embedding.}
    \begin{tabular}{cccc}
    \toprule
    \textbf{Dataset} & \textbf{Env} & \textbf{DeMa with pos. embed.} & \textbf{DeMa without pos. embed.} \\
    \midrule
    M & HalfCheetah & 42.8±0 & \multicolumn{1}{c}{\textbf{43±0.01}} \\
    M & Hopper & 71.2±14.6 & \multicolumn{1}{c}{\textbf{74.5±2.9}} \\
    M & Walker & \textbf{77.2±0.1} & \multicolumn{1}{c}{76.6±0.2} \\
    \midrule
    M-R & HalfCheetah & 40.2±0.1 & \multicolumn{1}{c}{\textbf{40.7±0.03}} \\
    M-R & Hopper & 77.2±35 & \multicolumn{1}{c}{\textbf{90.7±6.1}} \\
    M-R & Walker & 69.1±10.2 & \multicolumn{1}{c}{\textbf{70.5±0.1}} \\
    \midrule
    \multicolumn{2}{c}{\textbf{Average}} & 63.0 & \multicolumn{1}{c}{\textbf{66.0 }} \\
    \midrule
    \multicolumn{2}{c}{\textbf{All params \#}} & 431.5K & \multicolumn{1}{c}{\textbf{175.5K}} \\
    \bottomrule
    \end{tabular}%
  \label{tab:time embedding}%
\end{table}%

Position embedding is generally used in transformers to help the model understand the sequential nature of the data. It's a way of encoding the position of tokens in the sequence, and it can be crucial in tasks where the order of the data matters. Although we can use DeMa similar to using a transformer, which has input and output dimensions of (B, L, D) during training and inference, it differs in that it does not require position embedding to help the model have the ability to remember sequential information. As shown in Table~\ref{tab:time embedding}, the addition of position embedding not only failed to enhance the performance of the model but also led to a significant decrease in performance on certain tasks. Additionally, the introduction of position embedding significantly increased the model's parameter count, thereby adding to its computational burden. This finding highlights the advantage of the DeMa in terms of lightweight design, indicating its suitability for tasks with limited resources.

\section{Integrating DeMa with Other Methods}

We conduct additional experiments to show that DeMa can be combined with other trajectory optimization methods to achieve even better performance. By integrating DeMa with QT~\cite{QT}, we develop Q-DeMa. As shown in Table \ref{tab:qt}, Q-DeMa achieves performance comparable to state-of-the-art models while utilizing less than one-seventh of the parameter size of QT. This finding underscores the significant potential of applying Mamba to RL.

\begin{table}[htbp]
  \centering
  \caption{Q-DeMa's Results for D4RL datasets.}
    \begin{tabular}{cccccc}
    \toprule
    \textbf{Dataset} & \textbf{Env-Gym} & \textbf{DT} & 
    \textbf{DeMa(Ours)} & \textbf{QT} & \textbf{Q-DeMa(Ours)} \\
    \midrule
    M & HalfCheetah & 42.6  & 43±0.01 & 51.4 ± 0.4 & 51.2±0.04 \\
    M & Hopper & 68.4  & 74.5±2.9 & 96.9 ± 3.1 & 88.1±9.61 \\
    M & Walker & 75.5  & 76.6±0.2 & 88.8 ± 0.5 & 89.1±0.2 \\
    \midrule
    M-R & HalfCheetah & 37.0  & 40.7±0.03 & 48.9 ± 0.3 & 48.6±0.3 \\
    M-R & Hopper & 85.6  & 90.7±6.1 & 102.0 ± 0.2 & 101.5±0.1 \\
    M-R & Walker & 71.2  & 70.5±0.1 & 98.5 ± 1.1 & 99.8±1 \\
    \midrule
    \multicolumn{2}{c}{\textbf{Average-Gym}} & 63.4  & 66.0 & 81.0 & 79.7  \\
    \midrule
    \multicolumn{2}{c}{\textbf{All params \#}} & 726.2K/2.6M & 175.5K/500.0K & 3.7M & 500K \\
    \bottomrule
    \end{tabular}%
  \label{tab:qt}%
\end{table}%

\section{MuJoCo and Atari Tasks Scores}

Table~\ref{tab:baseline score} shows the normalized scores used in MuJoCo and Atari tasks, followed by~\cite{d4rl} and ~\cite{ye2021mastering}.

\begin{table}[htbp]
  \centering
  \caption{MuJoCo and Atari baseline scores used for normalization}
    \begin{tabular}{cccc}
    \toprule
    &\textbf{Env/Game} & \textbf{Random} & \textbf{Expert/Gamer} \\
    \midrule
     &Hopper & -20.3 & 3234.3 \\
    Gym&Halfcheetah & -280.2 & 12135 \\
    &Walker2d & 1.6 & 4592.3 \\
    \midrule
    &Breakout & 1.7 & 30.5 \\
   &Qbert & 163.9 & 13455 \\
    &Pong & -20.7 & 14.6 \\
    Atari&Seaquest & 68.4 & 42054.7 \\
    &Asterix & 210 & 8503 \\
    &Frostbite & 65 & 4335 \\
    &Assault & 222 & 742 \\
    &Gopher & 258 & 2412 \\
    \bottomrule
    \end{tabular}%
  \label{tab:baseline score}%
\end{table}%


\section{Types of Hidde Attention Scores}\label{app:attention}

\begin{figure}[htbp]
    \centering
    \begin{subfigure}[b]{\textwidth}
    \centering
    \includegraphics[width=\linewidth]{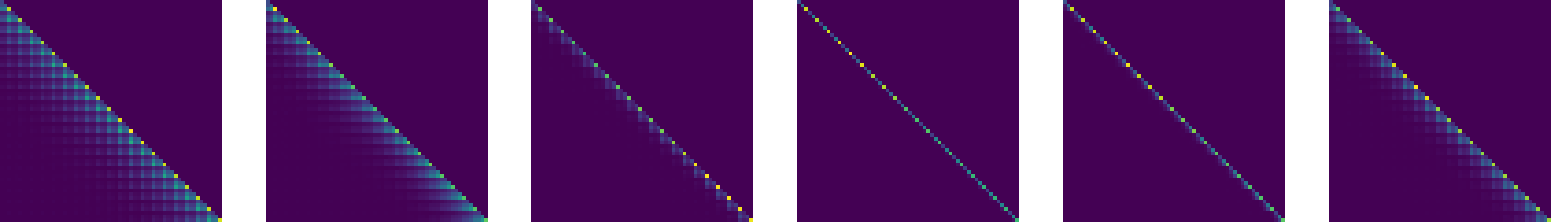}
    \caption{channel 30 to channel 36, layer 1}
    \end{subfigure}
    \begin{subfigure}[b]{\textwidth}
    \centering
    \includegraphics[width=\linewidth]{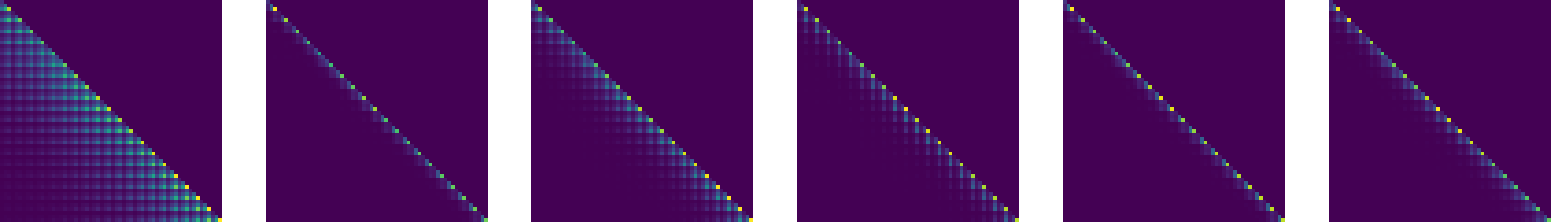}
    \caption{channel 30 to channel 36, layer 2}
    \end{subfigure}
    \begin{subfigure}[b]{\textwidth}
    \centering
    \includegraphics[width=\linewidth]{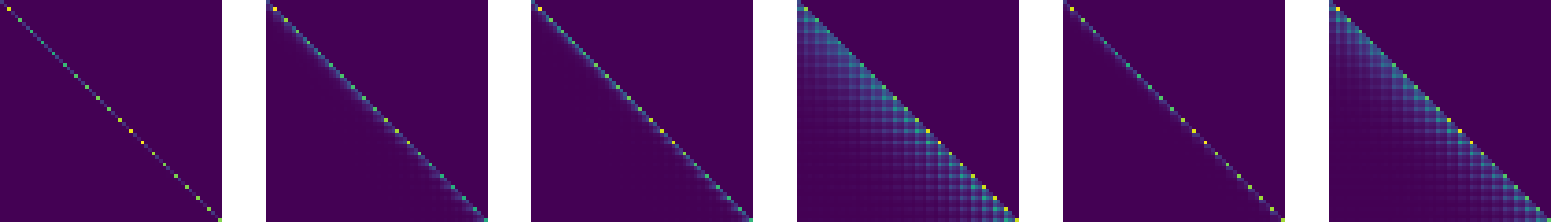}
    \caption{channel 30 to channel 36, layer 3}
    \end{subfigure}
    \hfill
    \begin{subfigure}[b]{.24\textwidth}
    \centering
    \includegraphics[width=\linewidth]{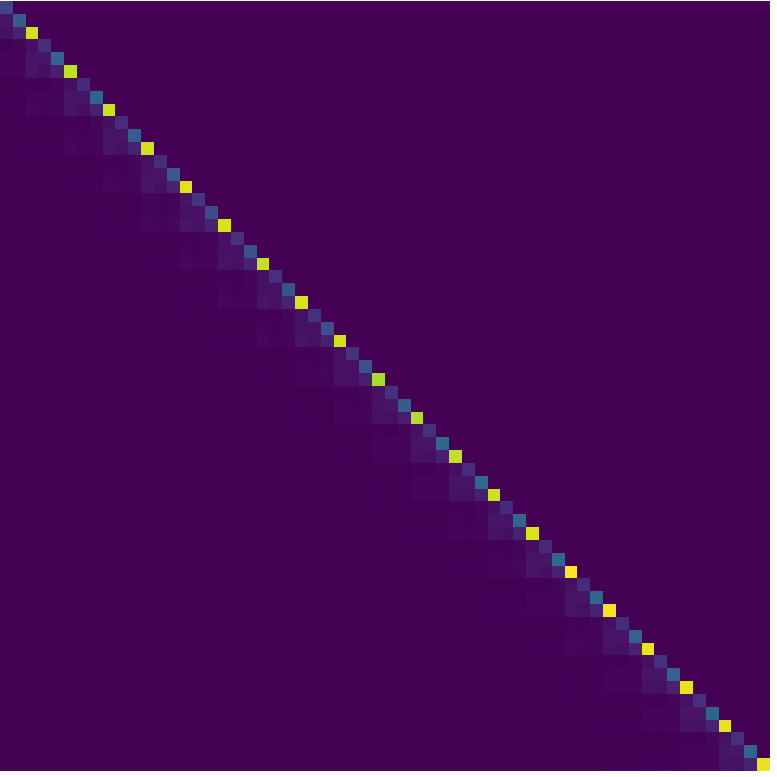}
    \caption{fused channel, layer 1}
    \end{subfigure}
    \hfill
    \begin{subfigure}[b]{.24\textwidth}
    \centering
    \includegraphics[width=\linewidth]{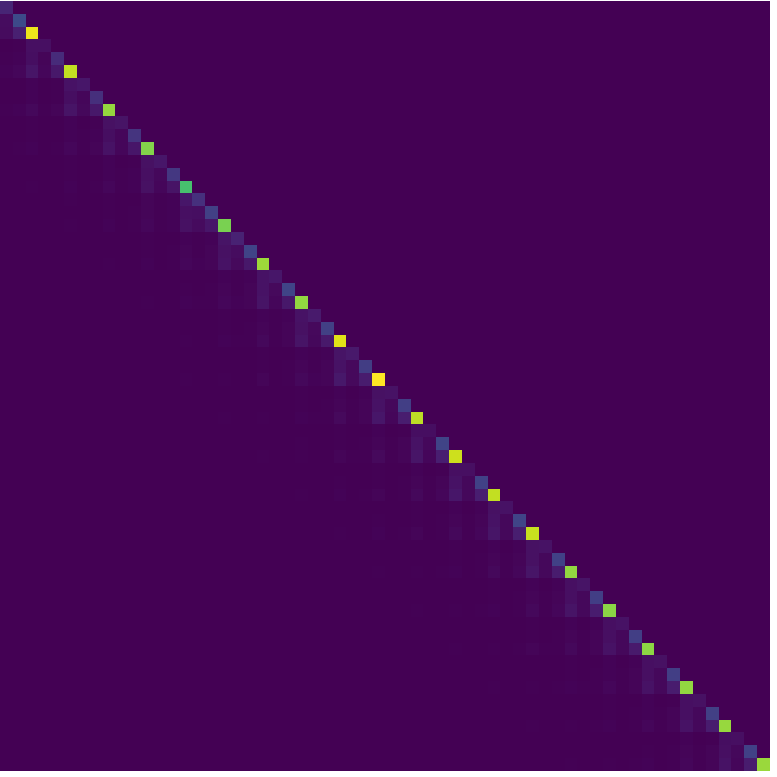}
    \caption{fused channel, layer 2}
    \end{subfigure}
    \hfill
    \begin{subfigure}[b]{.24\textwidth}
    \centering
    \includegraphics[width=\linewidth]{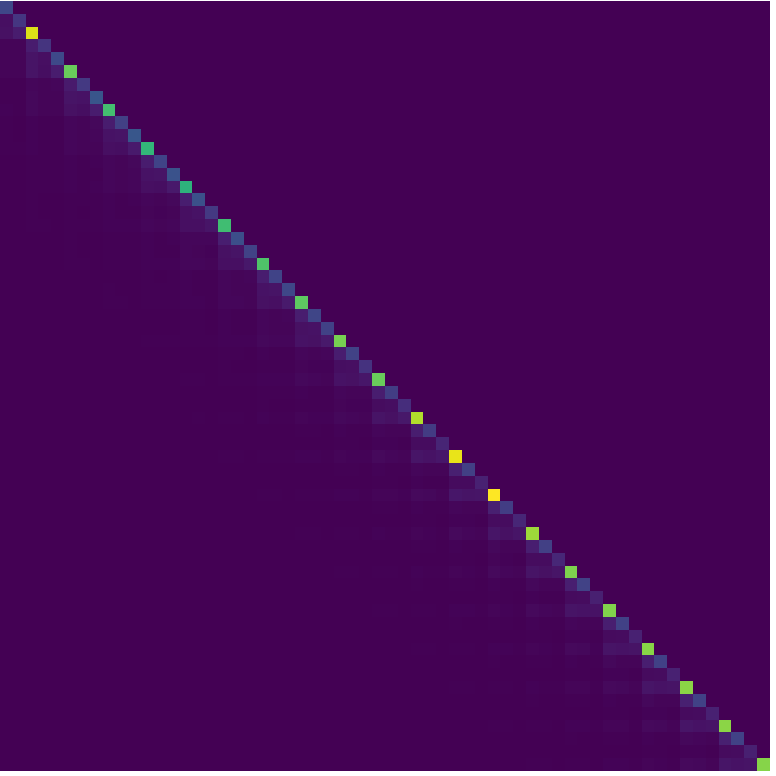}
    \caption{fused channel, layer 3}
    \end{subfigure}
    \hfill
    \begin{subfigure}[b]{.24\textwidth}
    \centering
    \includegraphics[width=\linewidth]{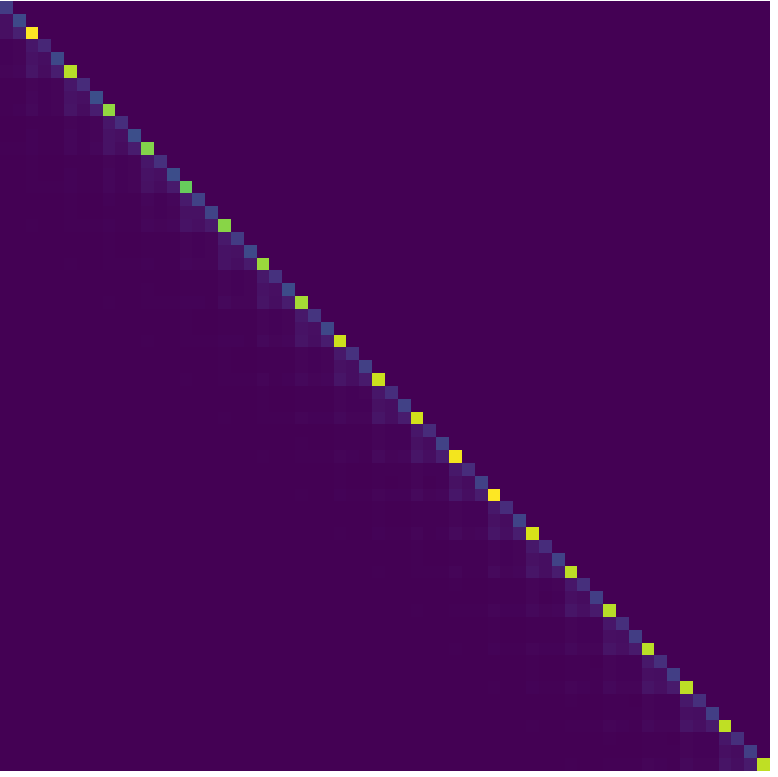}
    \caption{fused channel and layer}
    \end{subfigure}
    \caption{Hidden Attention Score Matrix of each channel and layer of DeMa, trained on the Hopper-medium dataset. The element $A_{ij}$ present the attention score between output $y_i$ and input $x_j$}
    \label{fig:下三角}
\end{figure}

\begin{figure}[htbp]
    \centering
    \begin{subfigure}[b]{0.34\textwidth}
    \centering
    \includegraphics[width=\linewidth]
    {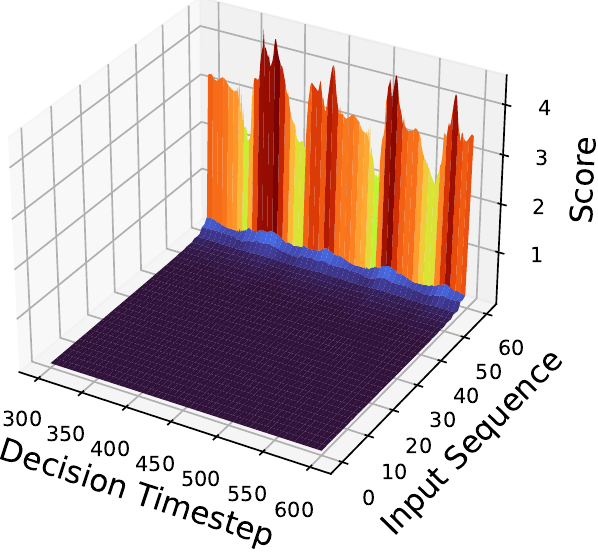}
    \subcaption{Layer 1}
    \end{subfigure}
    \hfill
    \begin{subfigure}[b]{0.34\textwidth}
    \centering
    \includegraphics[width=\linewidth]{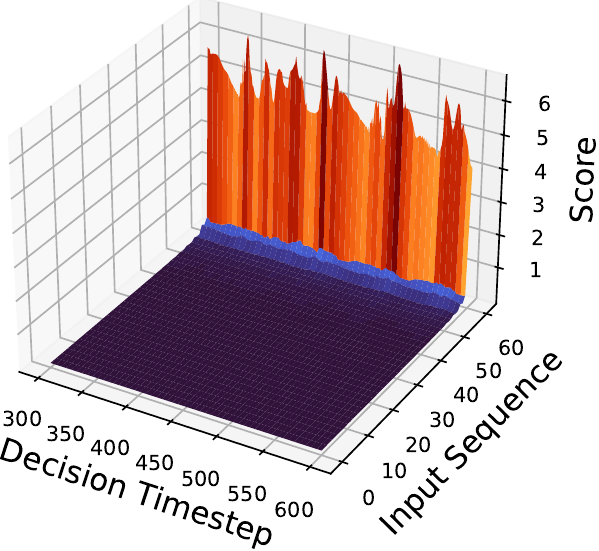}
    \subcaption{Layer 2}
    \end{subfigure}
    \hfill
    \begin{subfigure}[b]{0.34\textwidth}
    \centering
    \includegraphics[width=\linewidth]{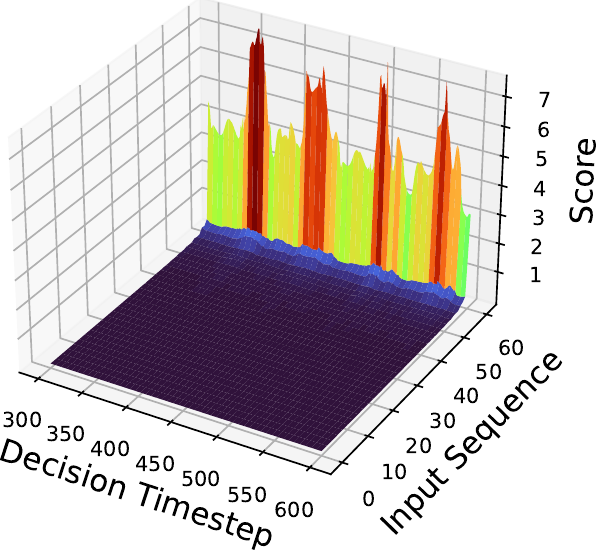}
    \subcaption{Layer 3}
    \end{subfigure}
    \hfill
    \begin{subfigure}[b]{0.34\textwidth}
    \centering
    \includegraphics[width=\linewidth]{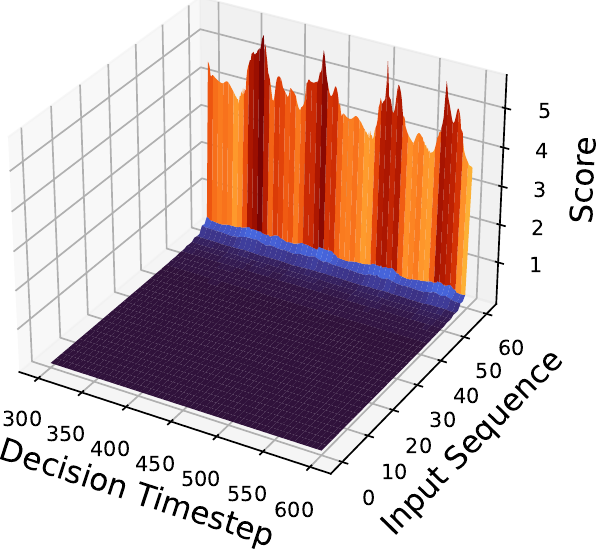}
    \subcaption{Fused Layer}
    \end{subfigure}
    \caption{\small  Hidden attention scores of DeMa from the 300th to the 600th timestep, trained on the Walker2d-medium dataset.}
    \label{fig:attention-mamba2}
    \vspace{-0.4cm}
\end{figure}

\begin{figure}[htbp]
    \centering
    \begin{subfigure}[b]{0.34\textwidth}
    \centering
    \includegraphics[width=\linewidth]
    {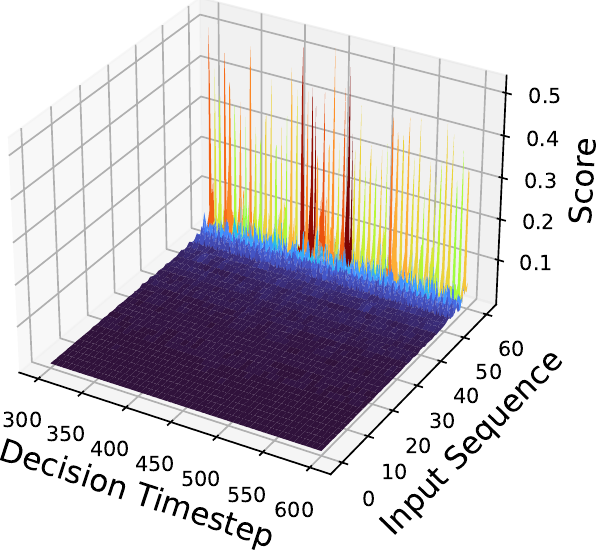}
    \subcaption{Layer 1}
    \end{subfigure}
    \hfill
    \begin{subfigure}[b]{0.34\textwidth}
    \centering
    \includegraphics[width=\linewidth]{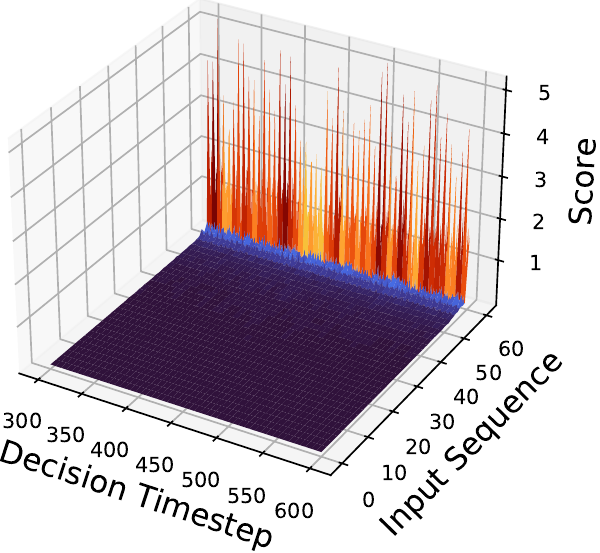}
    \subcaption{Layer 2}
    \end{subfigure}
    \hfill
    \begin{subfigure}[b]{0.34\textwidth}
    \centering
    \includegraphics[width=\linewidth]{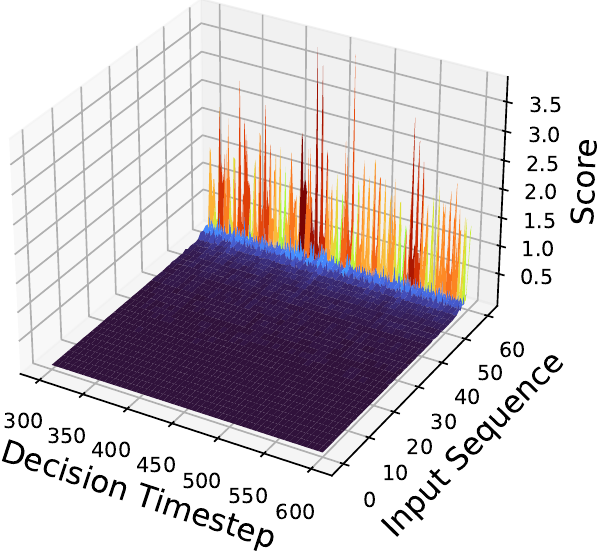}
    \subcaption{Layer 3}
    \end{subfigure}
    \hfill
    \begin{subfigure}[b]{0.34\textwidth}
    \centering
    \includegraphics[width=\linewidth]{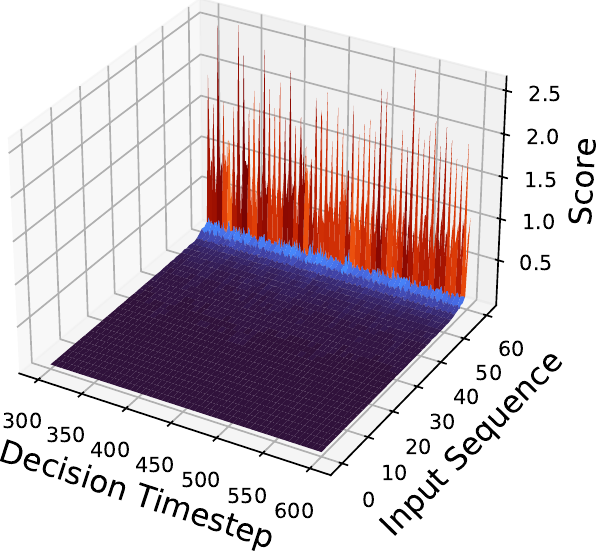}
    \subcaption{Fused Layer}
    \end{subfigure}
    \caption{\small  Hidden attention scores of DeMa from the 300th to the 600th timestep, trained on the Halfcheetah-medium-expert dataset.}
    \label{fig:attention-mamba3}
    \vspace{-0.4cm}
\end{figure}

\begin{figure}
\vspace{-1.8cm}
    \centering
    \begin{subfigure}[b]{0.24\textwidth}
    \centering
    \includegraphics[width=\linewidth]{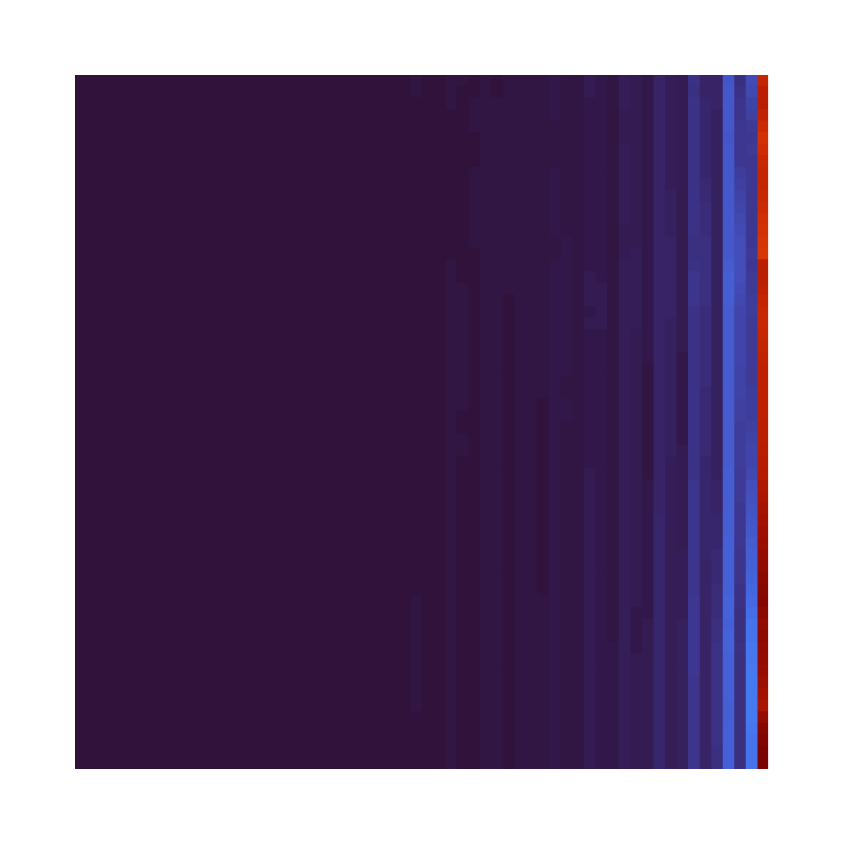}

    \end{subfigure}
    \begin{subfigure}[b]{0.24\textwidth}
    \centering
    \includegraphics[width=\linewidth]{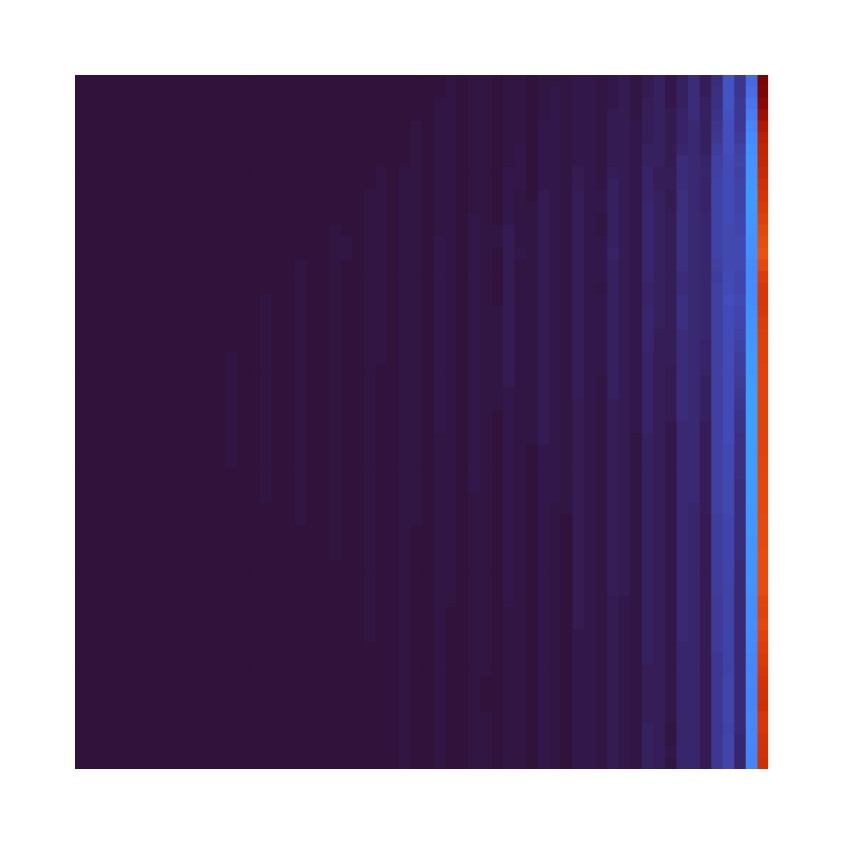}

    \end{subfigure}
    \begin{subfigure}[b]{0.24\textwidth}
    \centering
    \includegraphics[width=\linewidth]{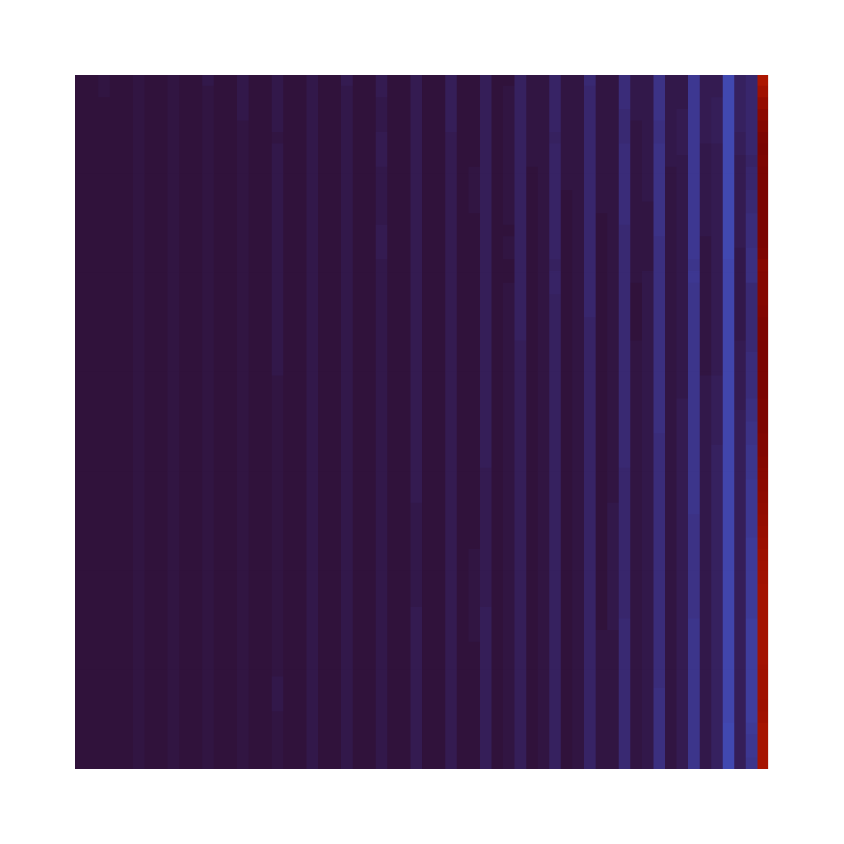}

    \end{subfigure}
    \begin{subfigure}[b]{0.24\textwidth}
    \centering
    \includegraphics[width=\linewidth]{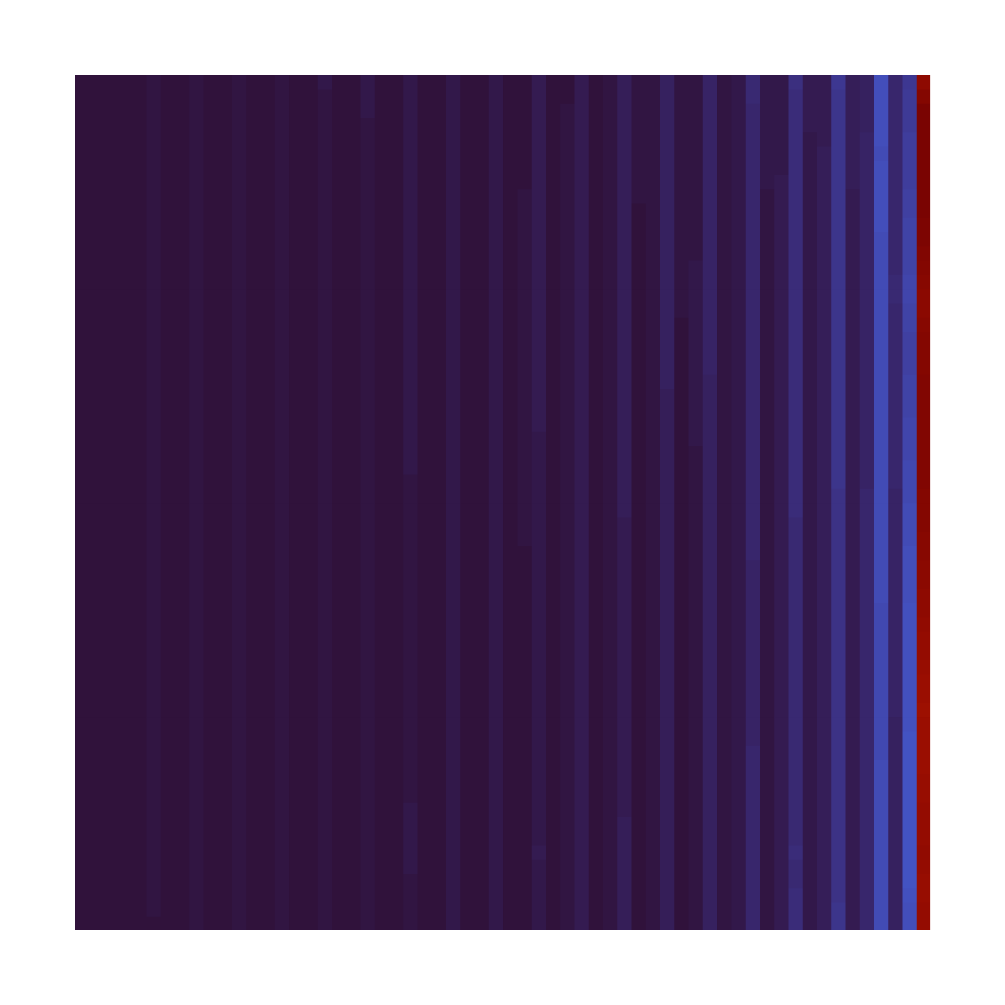}
    \end{subfigure}
    \\
    \begin{subfigure}[b]{0.24\textwidth}
    \centering
    \includegraphics[width=\linewidth]{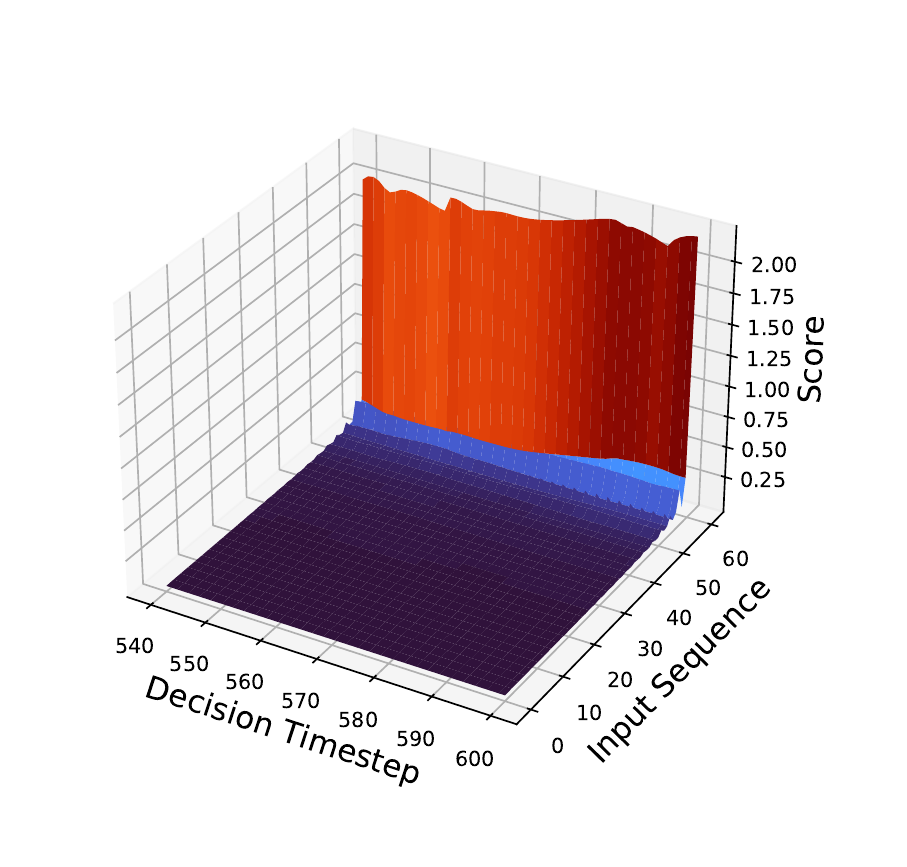}
    \subcaption{Layer 1}
    \end{subfigure}
    \begin{subfigure}[b]{0.24\textwidth}
    \centering
    \includegraphics[width=\linewidth]{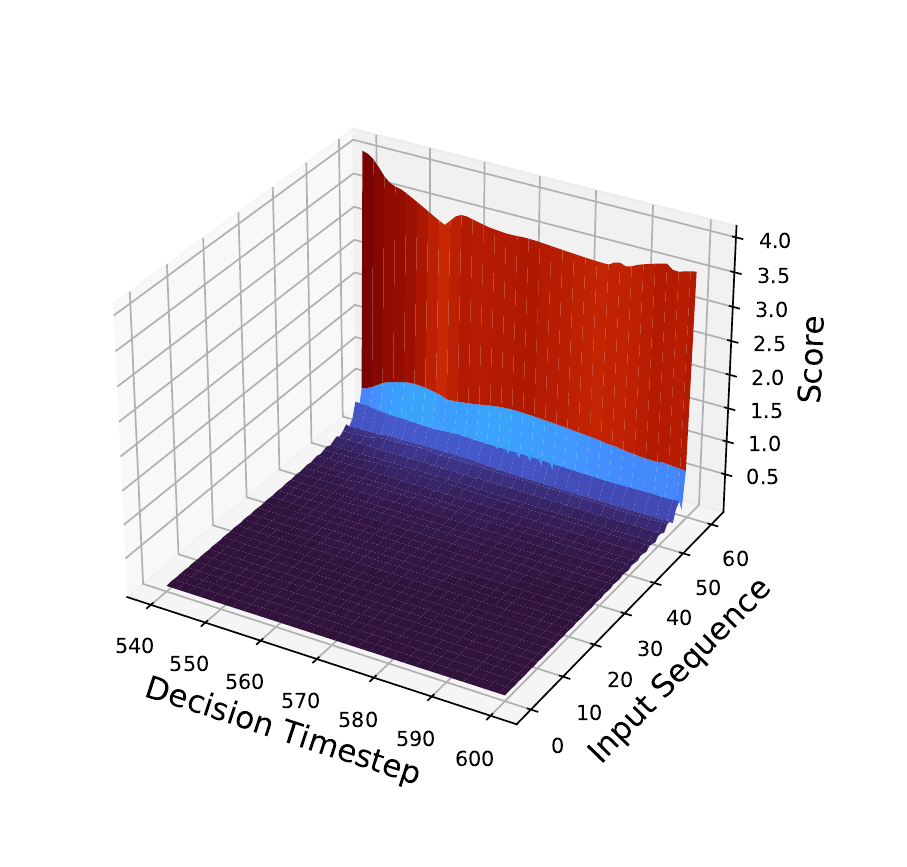}
    \subcaption{Layer 2}
    \end{subfigure}
    \begin{subfigure}[b]{0.24\textwidth}
    \centering
    \includegraphics[width=\linewidth]{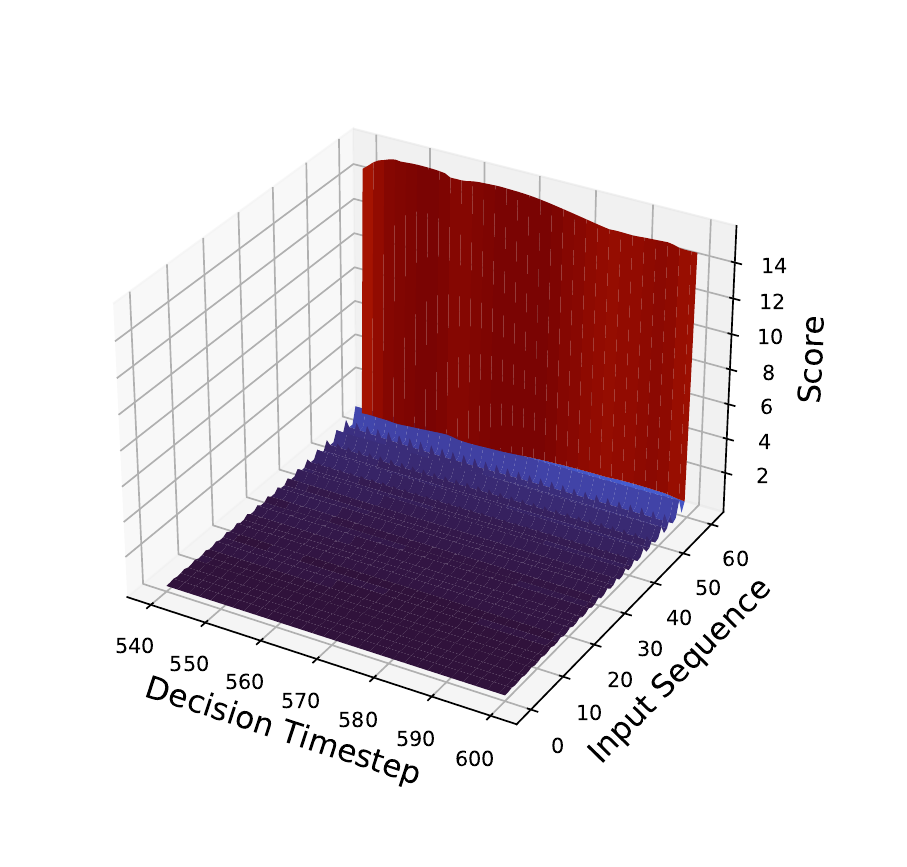}
    \subcaption{Layer 3}
    \end{subfigure}
    \begin{subfigure}[b]{0.24\textwidth}
    \centering
    \includegraphics[width=\linewidth]{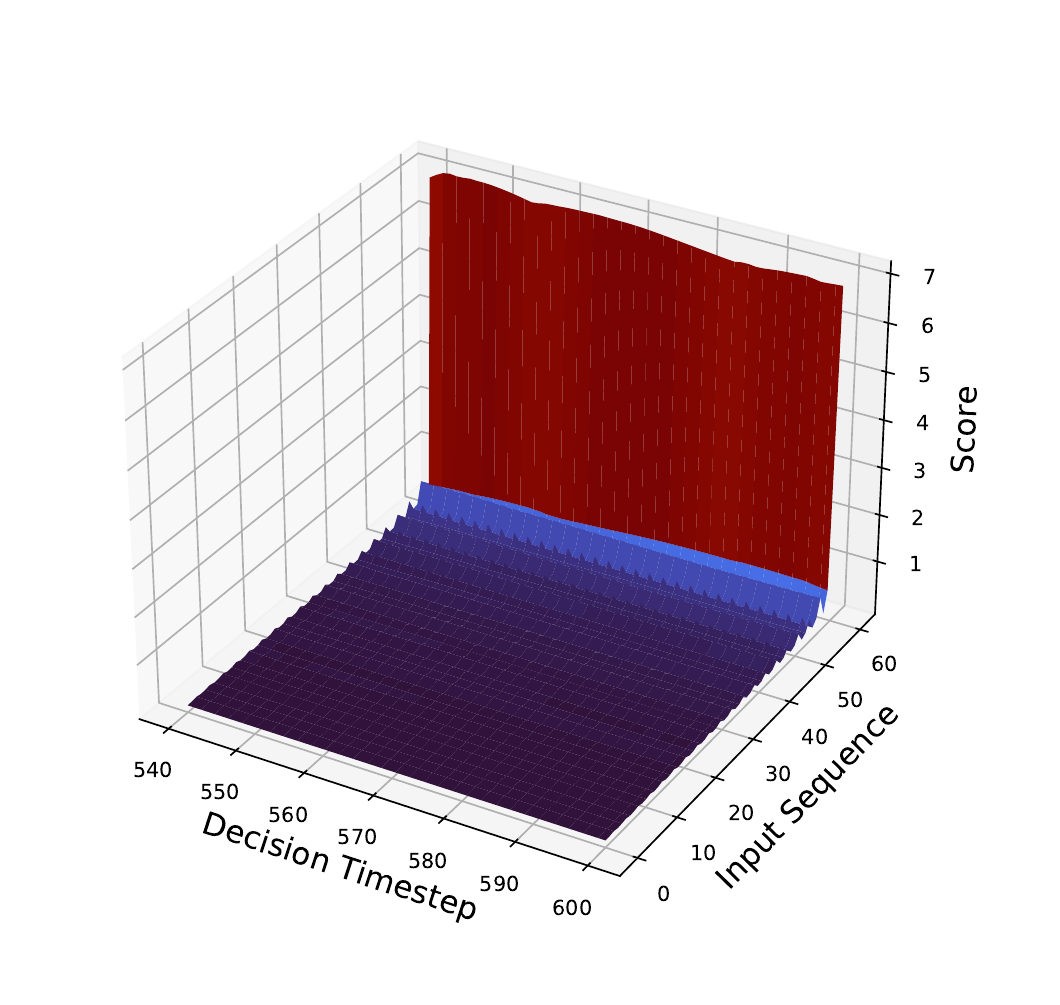}
    \subcaption{Fused Layer}
    \end{subfigure}
    \caption{\small  Hidden attention scores of DeMa from the 540th to the 600th timestep, trained on the Walker2d-medium dataset with delayed rewards. Hidden attention mechanism highlighting more focus on historical observations. \textbf{Top}: 2D Representation, \textbf{Bottom}: 3D Representation.}
    \label{fig:attention-mamba4}

    \centering
    \begin{subfigure}[b]{0.24\textwidth}
    \centering
    \includegraphics[width=\linewidth]{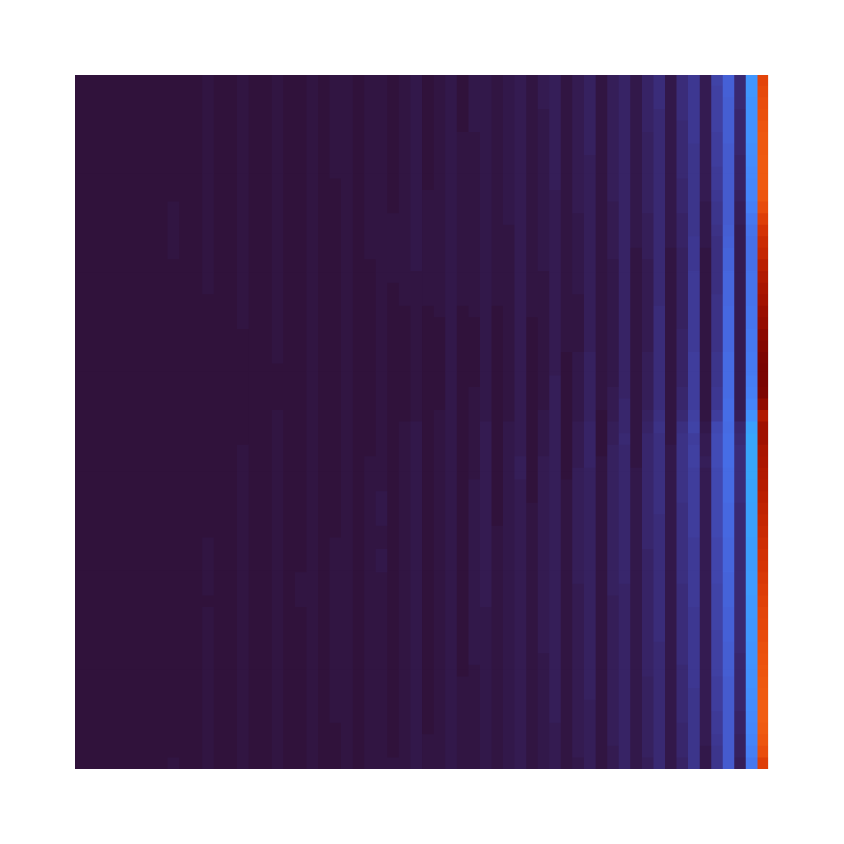}
    \end{subfigure}
    \begin{subfigure}[b]{0.24\textwidth}
    \centering
    \includegraphics[width=\linewidth]{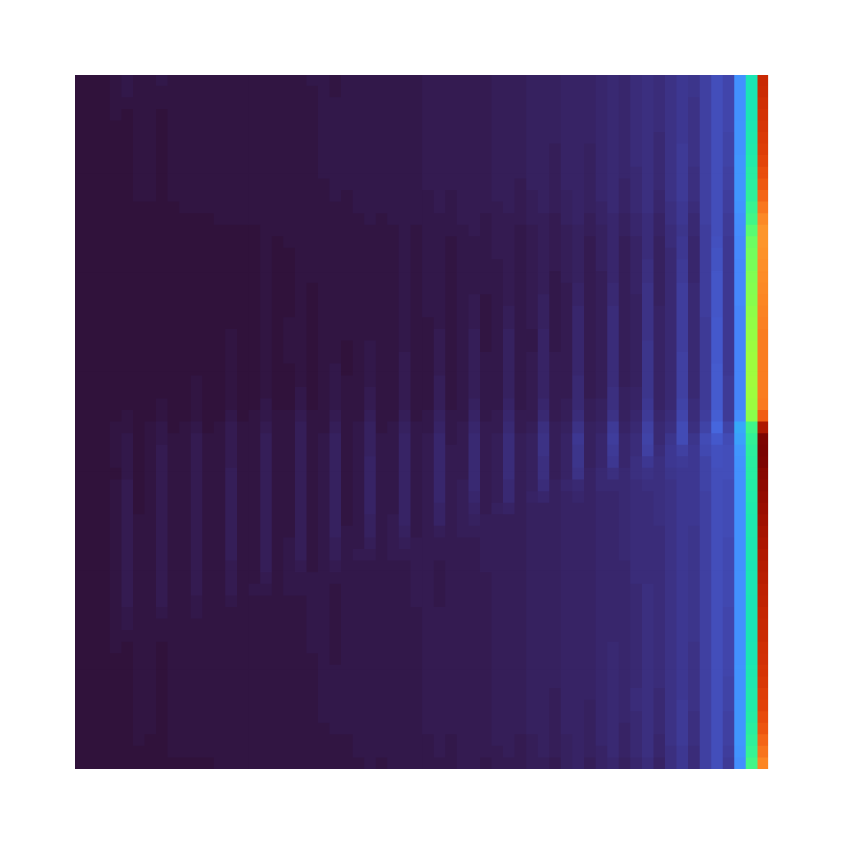}
    \end{subfigure}
    \begin{subfigure}[b]{0.24\textwidth}
    \centering
    \includegraphics[width=\linewidth]{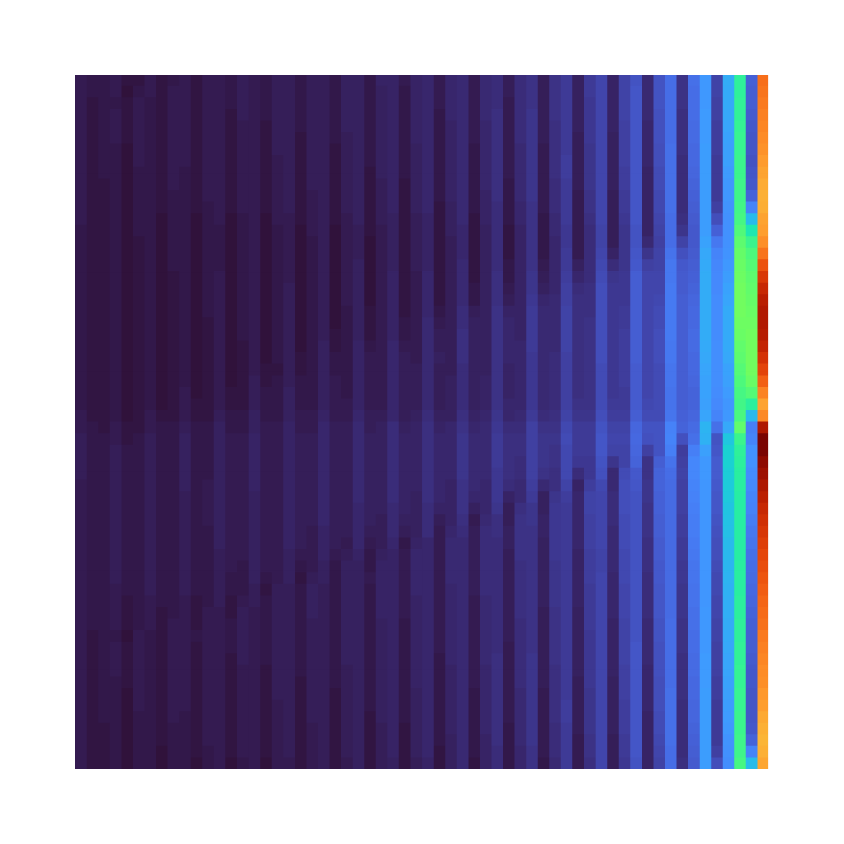}
    \end{subfigure}
    \begin{subfigure}[b]{0.24\textwidth}
    \centering
    \includegraphics[width=\linewidth]{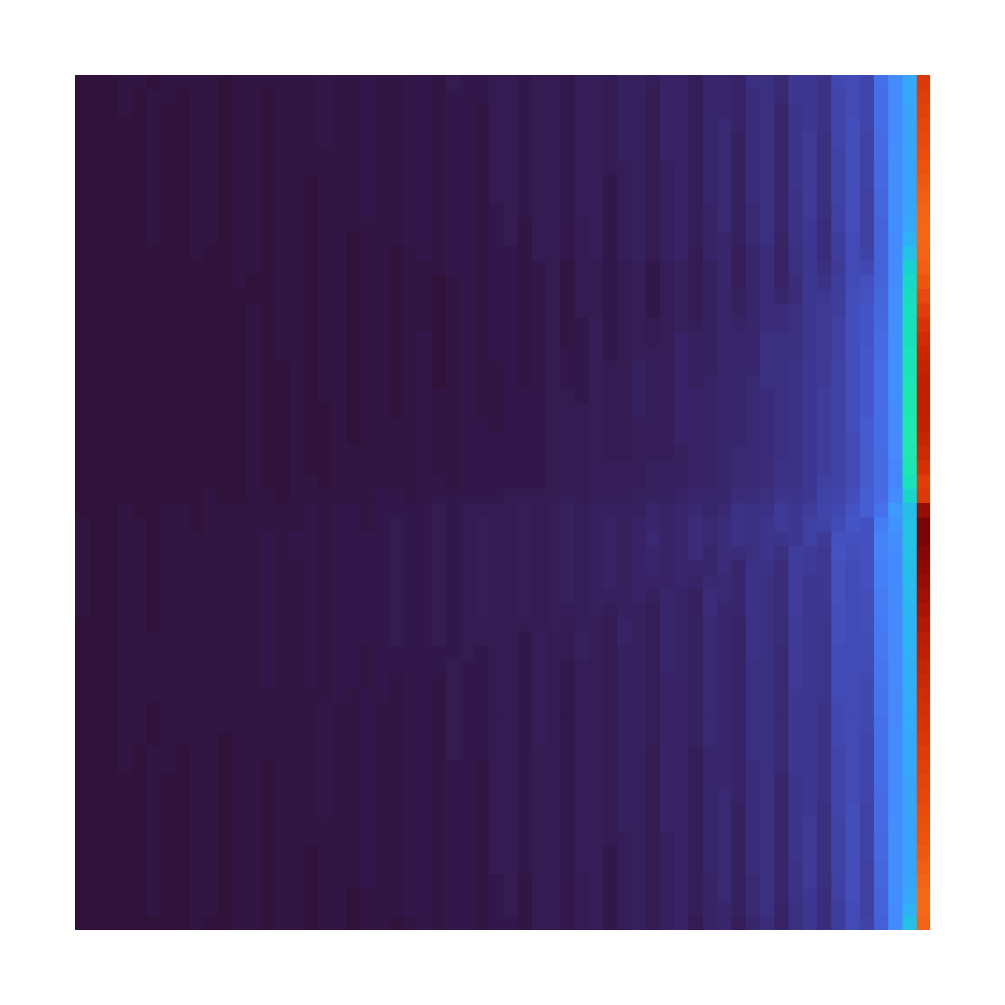}
    \end{subfigure}
    \\
    \begin{subfigure}[b]{0.24\textwidth}
    \centering
    \includegraphics[width=\linewidth]{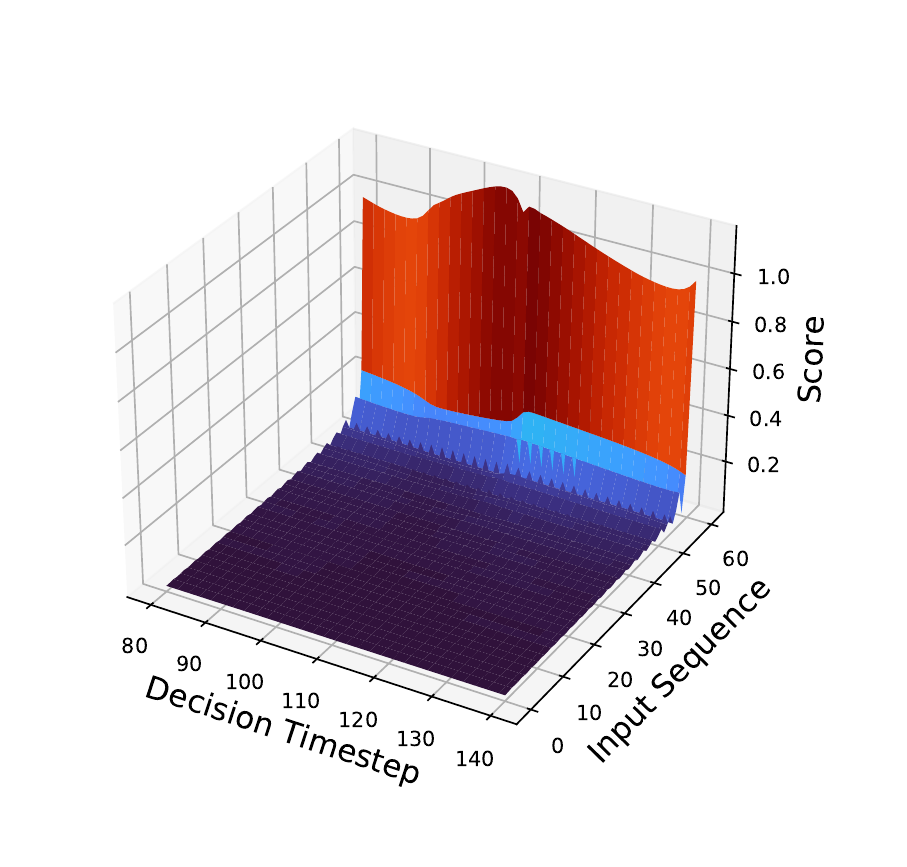}
    \subcaption{Layer 1}
    \end{subfigure}
    \begin{subfigure}[b]{0.24\textwidth}
    \centering
    \includegraphics[width=\linewidth]{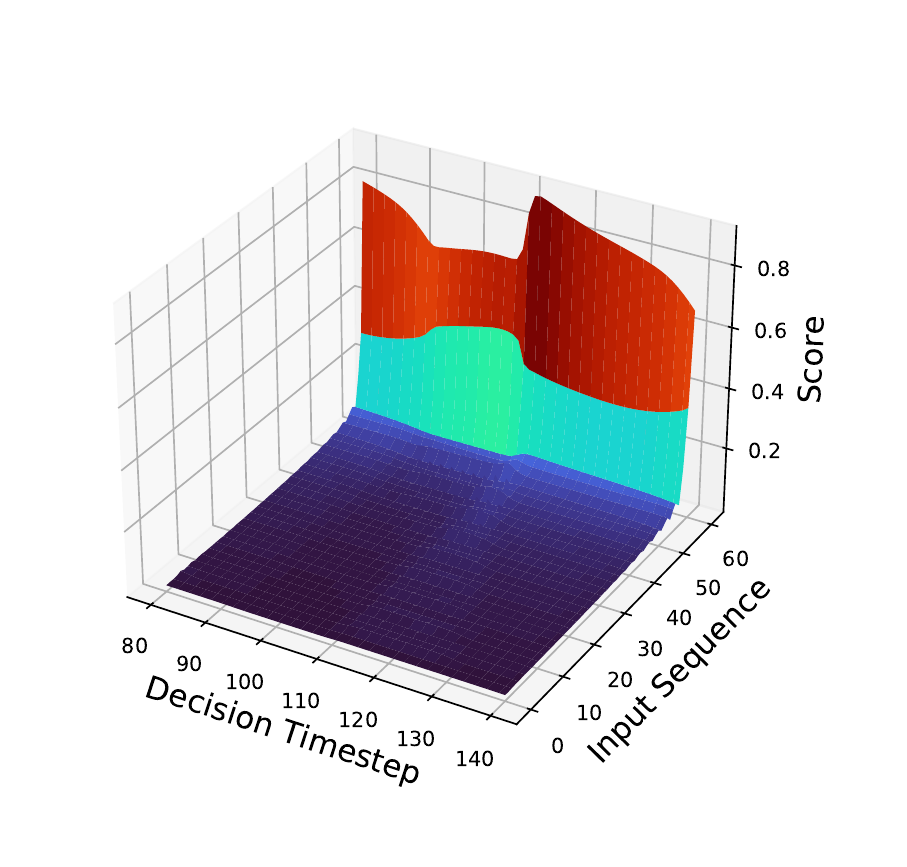}
    \subcaption{Layer 2}
    \end{subfigure}
    \begin{subfigure}[b]{0.24\textwidth}
    \centering
    \includegraphics[width=\linewidth]{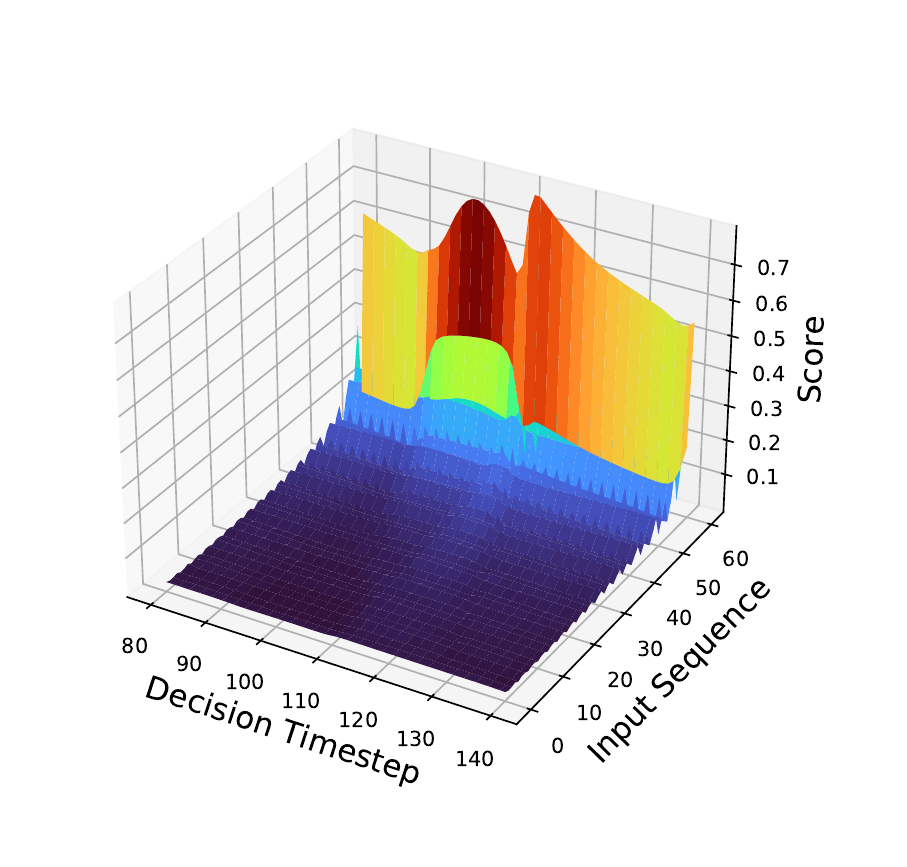}
    \subcaption{Layer 3}
    \end{subfigure}
    \begin{subfigure}[b]{0.24\textwidth}
    \centering
    \includegraphics[width=\linewidth]{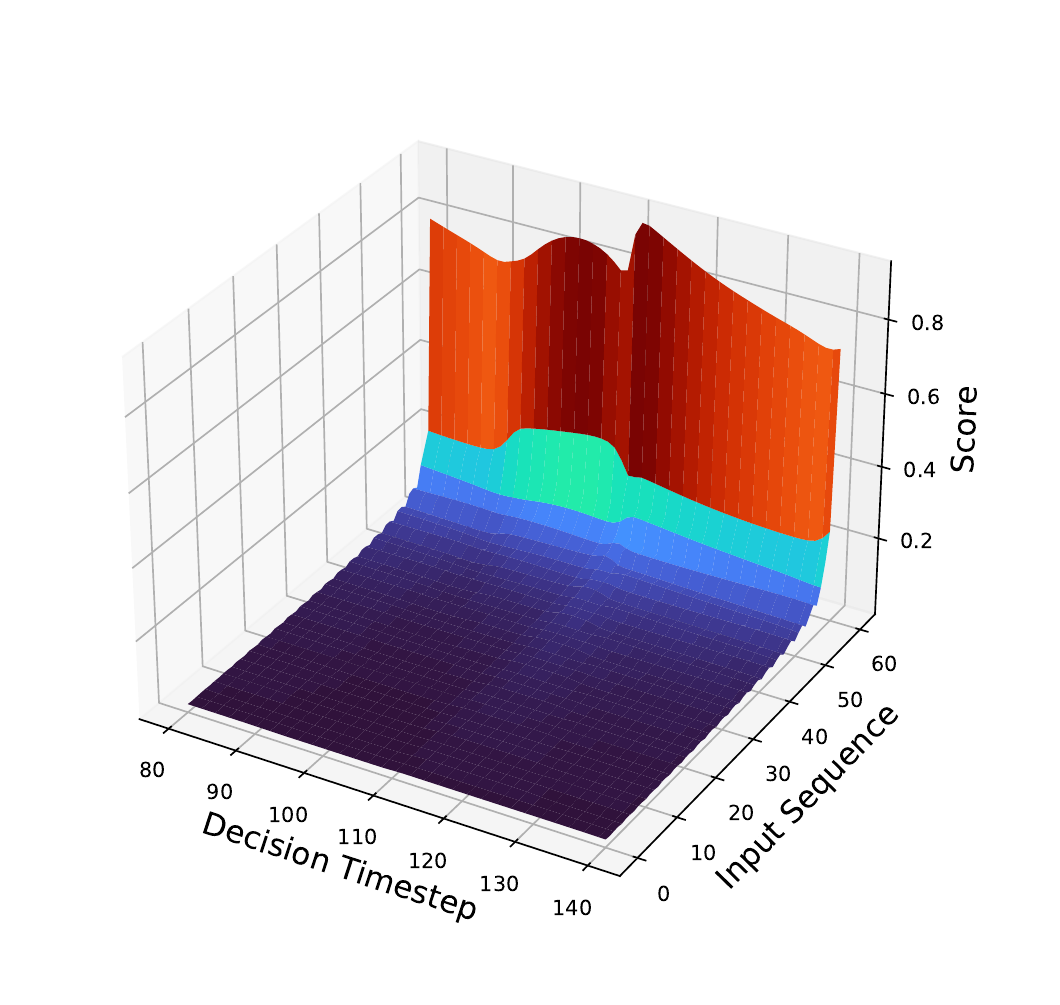}
    \subcaption{Fused Layer}
    \end{subfigure}
    \caption{\small  Hidden attention scores of DeMa from the 80th to the 140th timestep in maze2d-umaze. \textbf{Top}: 2D Representation, \textbf{Bottom}: 3D Representation.}
\label{fig:attention-mamba5}
    \centering
    \begin{subfigure}[b]{0.24\textwidth}
    \centering
    \includegraphics[width=\linewidth]{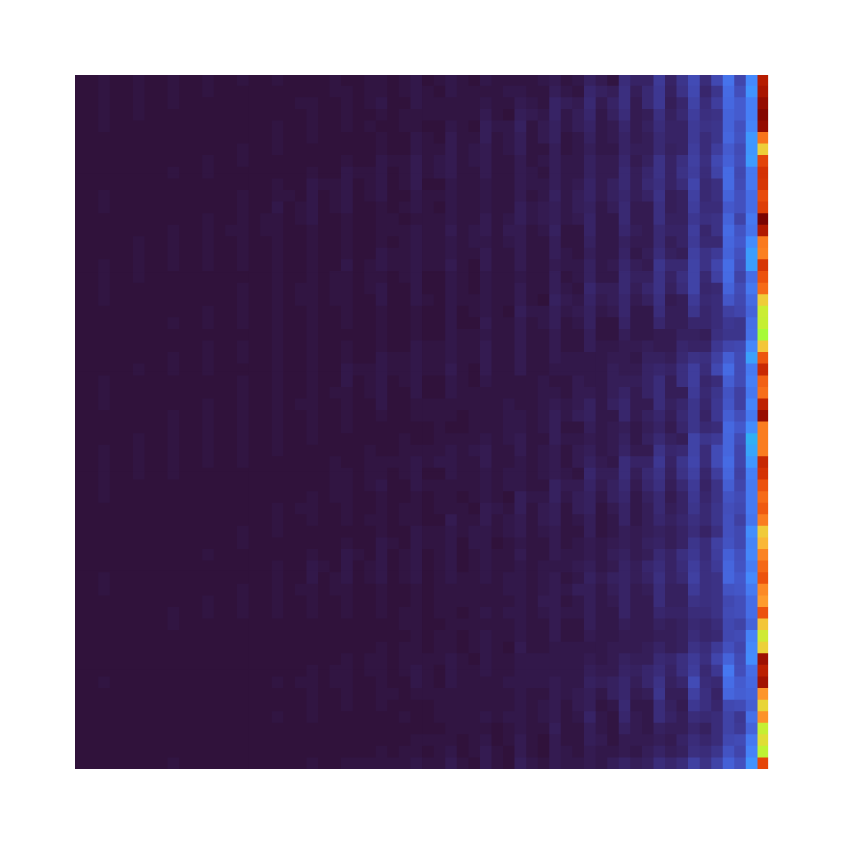}
    \end{subfigure}
    \begin{subfigure}[b]{0.24\textwidth}
    \centering
    \includegraphics[width=\linewidth]{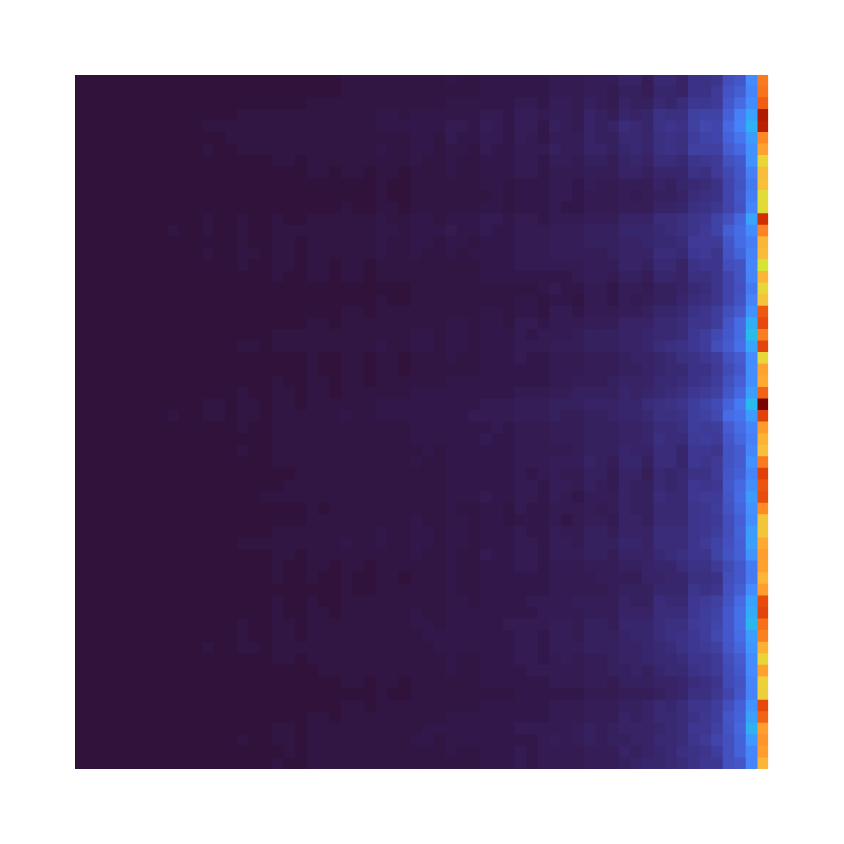}
    \end{subfigure}
    \begin{subfigure}[b]{0.24\textwidth}
    \centering
    \includegraphics[width=\linewidth]{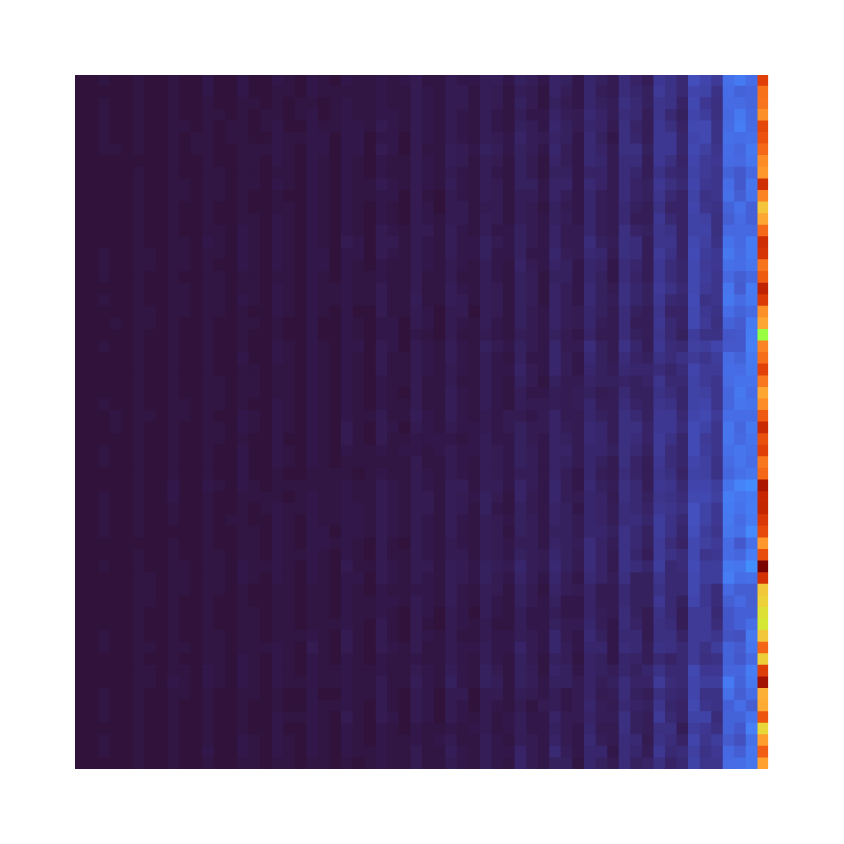}
    \end{subfigure}
    \begin{subfigure}[b]{0.24\textwidth}
    \centering
    \includegraphics[width=\linewidth]{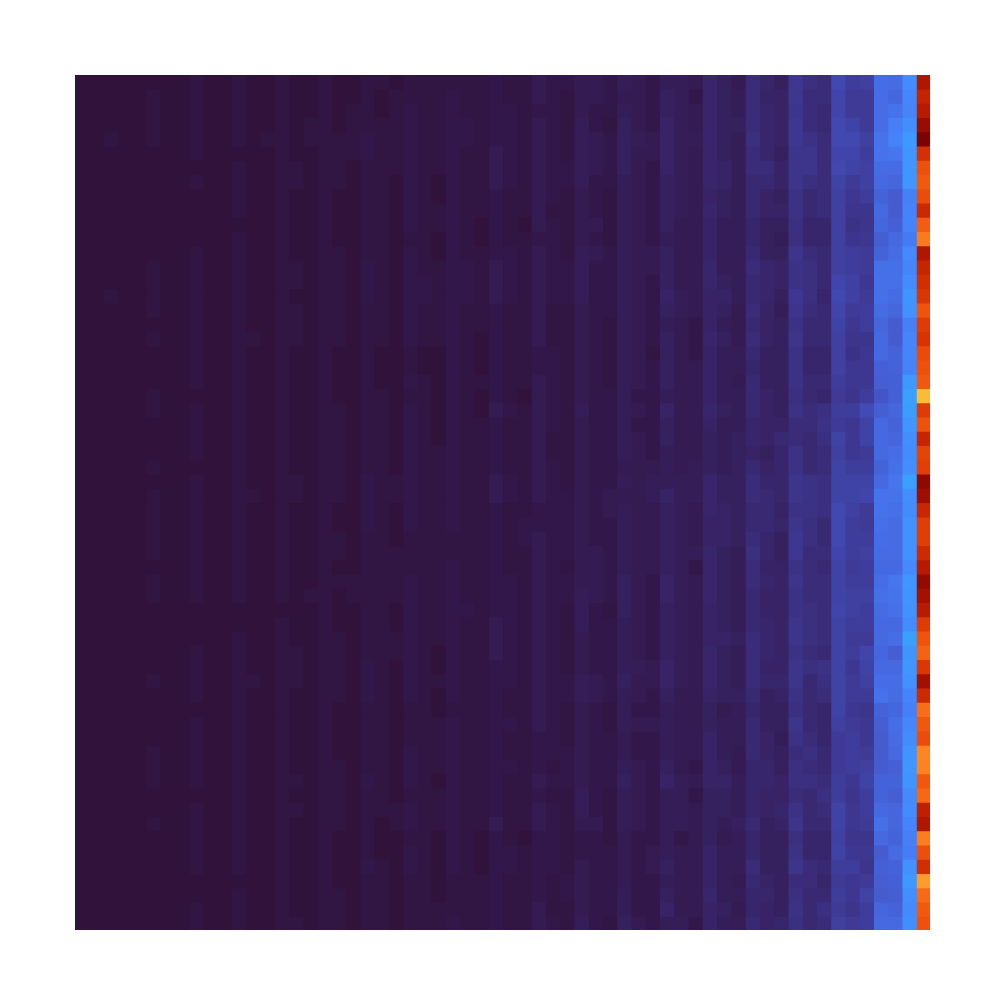}
    \end{subfigure}
    \\
    \begin{subfigure}[b]{0.24\textwidth}
    \centering
    \includegraphics[width=\linewidth]{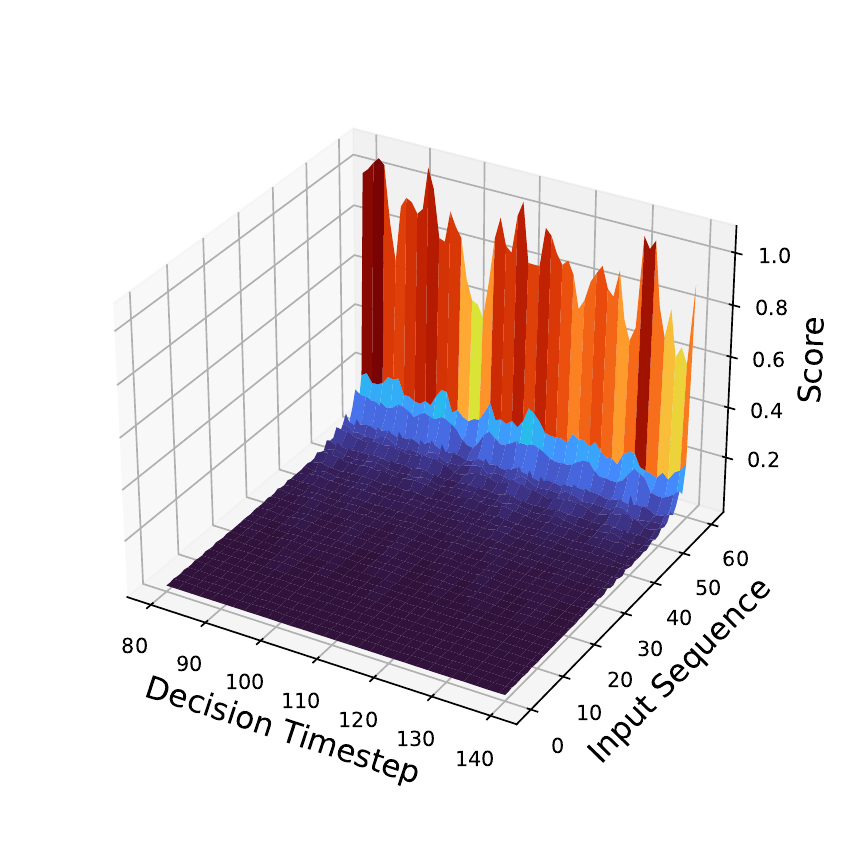}
    \subcaption{Layer 1}
    \end{subfigure}
    \begin{subfigure}[b]{0.24\textwidth}
    \centering
    \includegraphics[width=\linewidth]{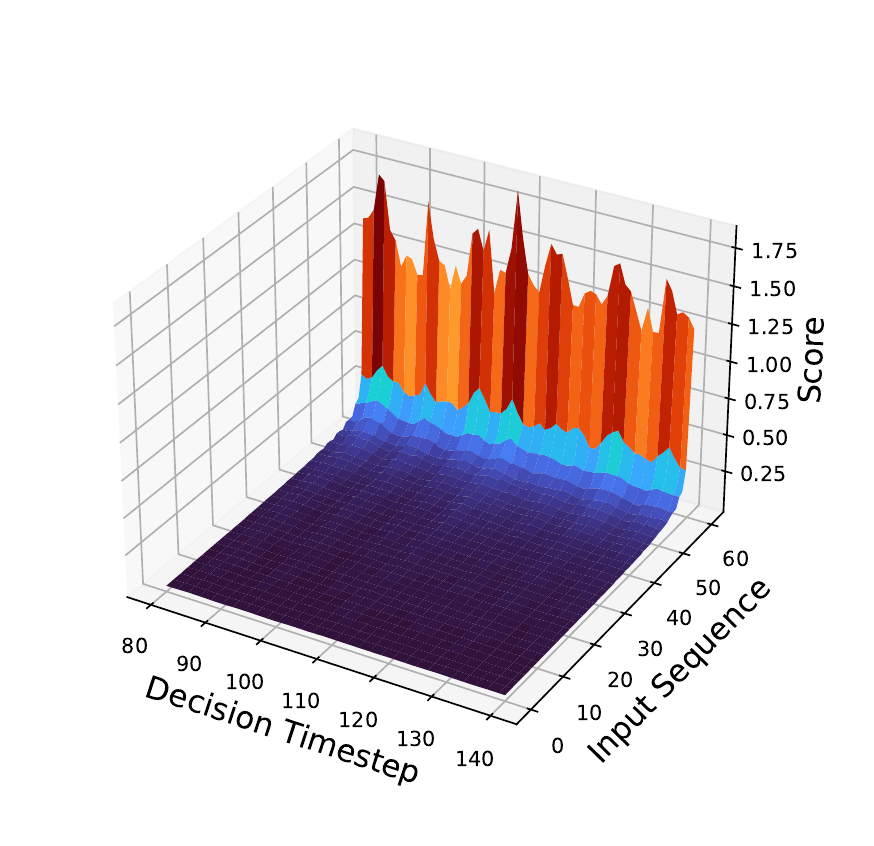}
    \subcaption{Layer 2}
    \end{subfigure}
    \begin{subfigure}[b]{0.24\textwidth}
    \centering
    \includegraphics[width=\linewidth]{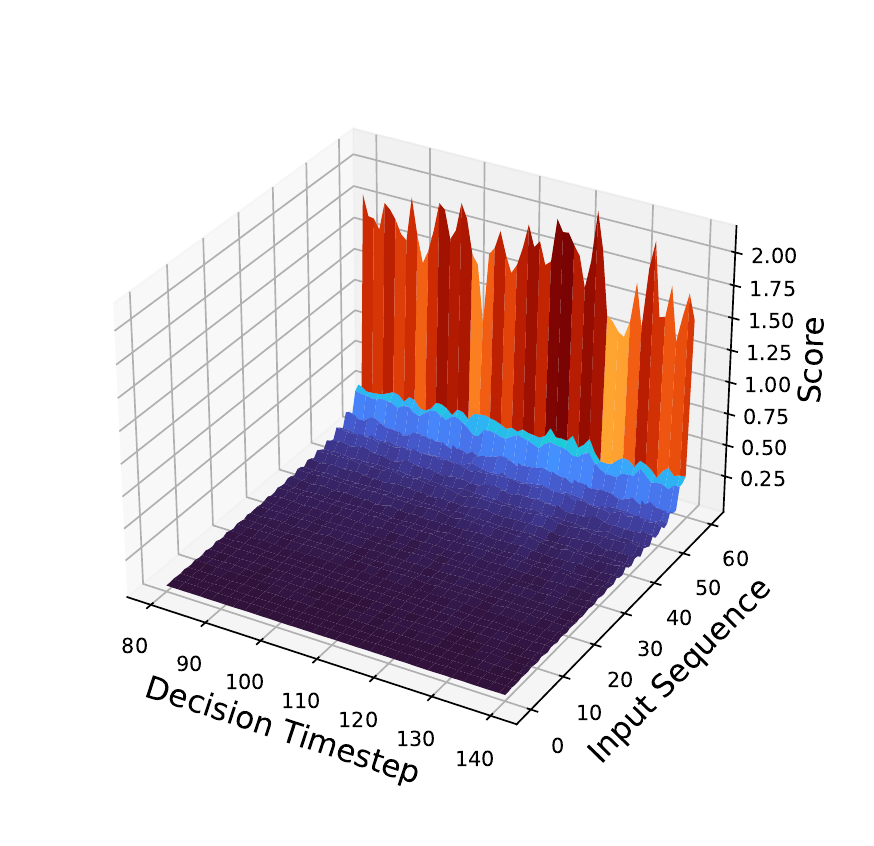}
    \subcaption{Layer 3}
    \end{subfigure}
    \begin{subfigure}[b]{0.24\textwidth}
    \centering
    \includegraphics[width=\linewidth]{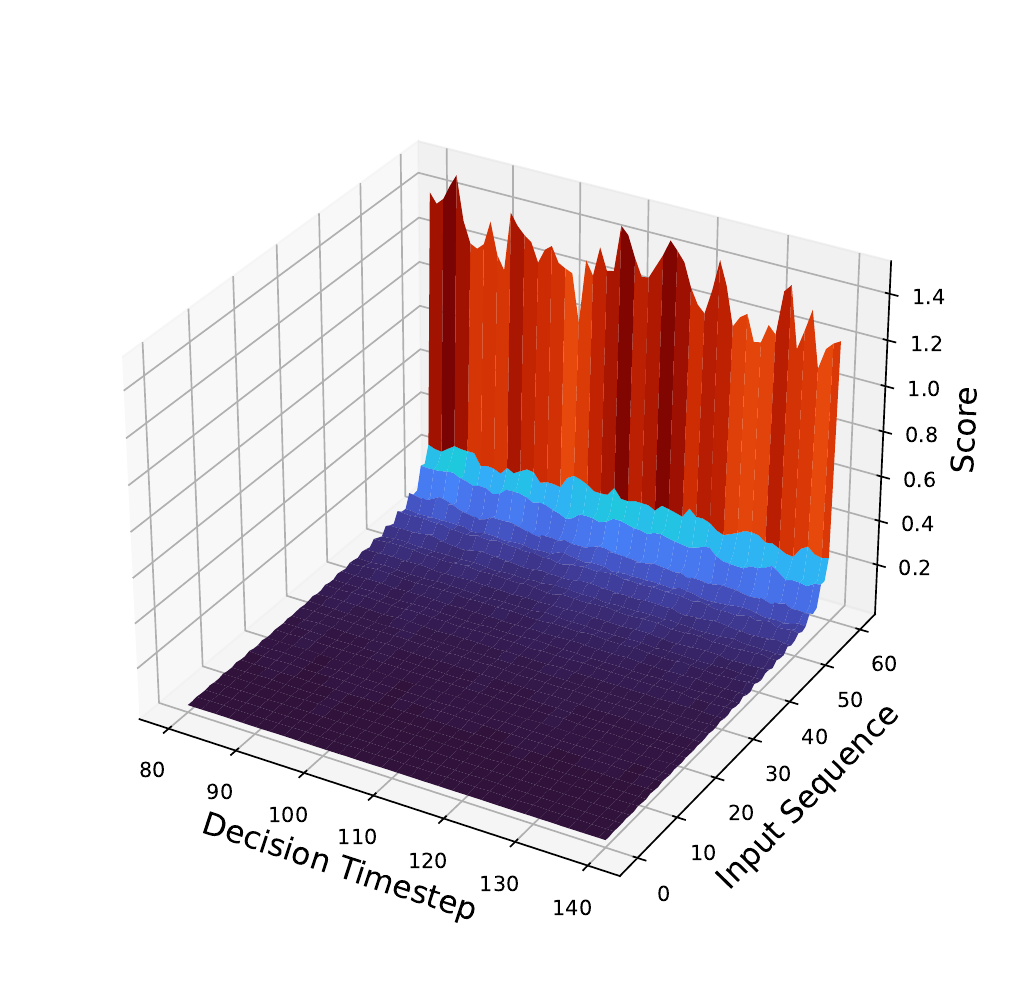}
    \subcaption{Fused Layer}
    \end{subfigure}
    \caption{\small  Hidden attention scores of DeMa from the 80th to the 140th timestep in antmaze-umaze-diverse. \textbf{Top}: 2D Representation, \textbf{Bottom}: 3D Representation.}
\label{fig:attention-mamba6}
\end{figure}

In the previous articles~\cite{dc, zhou2023adaptive}, the visualization of attention in DT was in the form of a lower triangular matrix. However, this lower-triangular matrix reflects the attention scores of each generated token to the input sequence during the training phase, and it cannot accurately illustrate the context information that the model focuses on at each decision-making step. As can be seen, the element in Figure~\ref{fig:下三角} at the i-th row and j-th column represents presents the output $y_i$'s attention score to input $x_j$. In the training phase, all corresponding predicting actions are used to calculate the reconstructed loss with target actions. However, during the evaluation phase, i.e. when interacting with the environment, we input $x:(1, L, D)$, and the model also outputs $y:(1, L, D)$. At this time, we only use the last one of the model's output, which is $y[:,-1,:]$, corresponding to the last row in the matrix. Therefore, it is not quite appropriate to judge the context information the model focuses on at each decision-making step based on the lower-triangular matrix in Figure~\ref{fig:下三角}, as we want to understand the model's decision-making behavior at each step, thus leads to the creation of Figure~\ref{fig:attention-mamba}, Figure~\ref{fig:attention-mamba2} and Figure~\ref{fig:attention-mamba3}. It also demonstrates strong forgetting characteristics. This aligns with the properties of Markov chains as described in ~\cite{dc}, where the sequence of states precisely forms a Markov chain. We also conduct additional explorations in environments involving delayed rewards(MuJoCo with delayed rewards) and maze navigation(maze2d, antmaze). The performance of the hidden attention mechanism is illustrated in Figures~\ref{fig:attention-mamba4}-\ref{fig:attention-mamba6}. Although DeMa’s attention to past information increases, the hidden attention mechanism still prioritizes the current information when the Markov property of the environment is relaxed. Furthermore, among historical information, the hidden attention mechanism demonstrates a significantly higher focus on states compared to rewards or actions.

\clearpage
\end{document}